\documentclass[10pt,twocolumn,letterpaper]{article}

\usepackage[pagenumbers]{wacv} 

\usepackage{graphicx}
\usepackage{amsmath}
\usepackage{amssymb}
\usepackage{booktabs} 

\usepackage{amsfonts}  
\usepackage{colortbl}  
\usepackage{comment}   
\usepackage{multirow}  
\usepackage{todonotes} 
\usepackage{enumitem}  
\usepackage[normalem]{ulem} %
\usepackage{xcolor}
\definecolor{darkgreen}{rgb}{0,0.65,0.31}

\usepackage[pagebackref,breaklinks,colorlinks]{hyperref}

\usepackage[capitalize]{cleveref}
\crefname{section}{Sec.}{Secs.}
\Crefname{section}{Section}{Sections}
\Crefname{table}{Table}{Tables}
\crefname{table}{Tab.}{Tabs.}

\def\wacvPaperID{1417} 
\def\confName{WACV}
\def\confYear{2024}

\begin{document}

\title{High-fidelity Pseudo-labels for Boosting Weakly-Supervised Segmentation}

\author{Arvi Jonnarth\thanks{Affiliation: Husqvarna Group, Huskvarna, Sweden.} \qquad Yushan Zhang \qquad Michael Felsberg\thanks{Co-affiliation: University of KwaZulu-Natal, Durban, South Africa.}\\
Department of Electrical Engineering, Linköping University, Sweden\\
{\tt\small \{arvi.jonnarth, yushan.zhang, michael.felsberg\}@liu.se}
}

\maketitle

\begin{abstract}
Image-level weakly-supervised semantic segmentation (WSSS) reduces the usually vast data annotation cost by surrogate segmentation masks during training. The typical approach involves training an image classification network using global average pooling (GAP) on convolutional feature maps. This enables the estimation of object locations based on class activation maps (CAMs), which identify the importance of image regions. The CAMs are then used to generate pseudo-labels, in the form of segmentation masks, to supervise a segmentation model in the absence of pixel-level ground truth. Our work is based on two techniques for improving CAMs; importance sampling, which is a substitute for GAP, and the feature similarity loss, which utilizes a heuristic that object contours almost always align with color edges in images. However, both are based on the multinomial posterior with softmax, and implicitly assume that classes are mutually exclusive, which turns out suboptimal in our experiments. Thus, we reformulate both techniques based on binomial posteriors of multiple independent binary problems. This has two benefits; their performance is improved and they become more general, resulting in an add-on method that can boost virtually any WSSS method. This is demonstrated on a wide variety of baselines on the PASCAL VOC dataset, improving the region similarity and contour quality of all implemented state-of-the-art methods. Experiments on the MS COCO dataset further show that our proposed add-on is well-suited for large-scale settings. Our code implementation is available at \url{https://github.com/arvijj/hfpl}.
\end{abstract}

\begin{figure}[t]
    \centering
    \includegraphics[width=0.79\columnwidth]{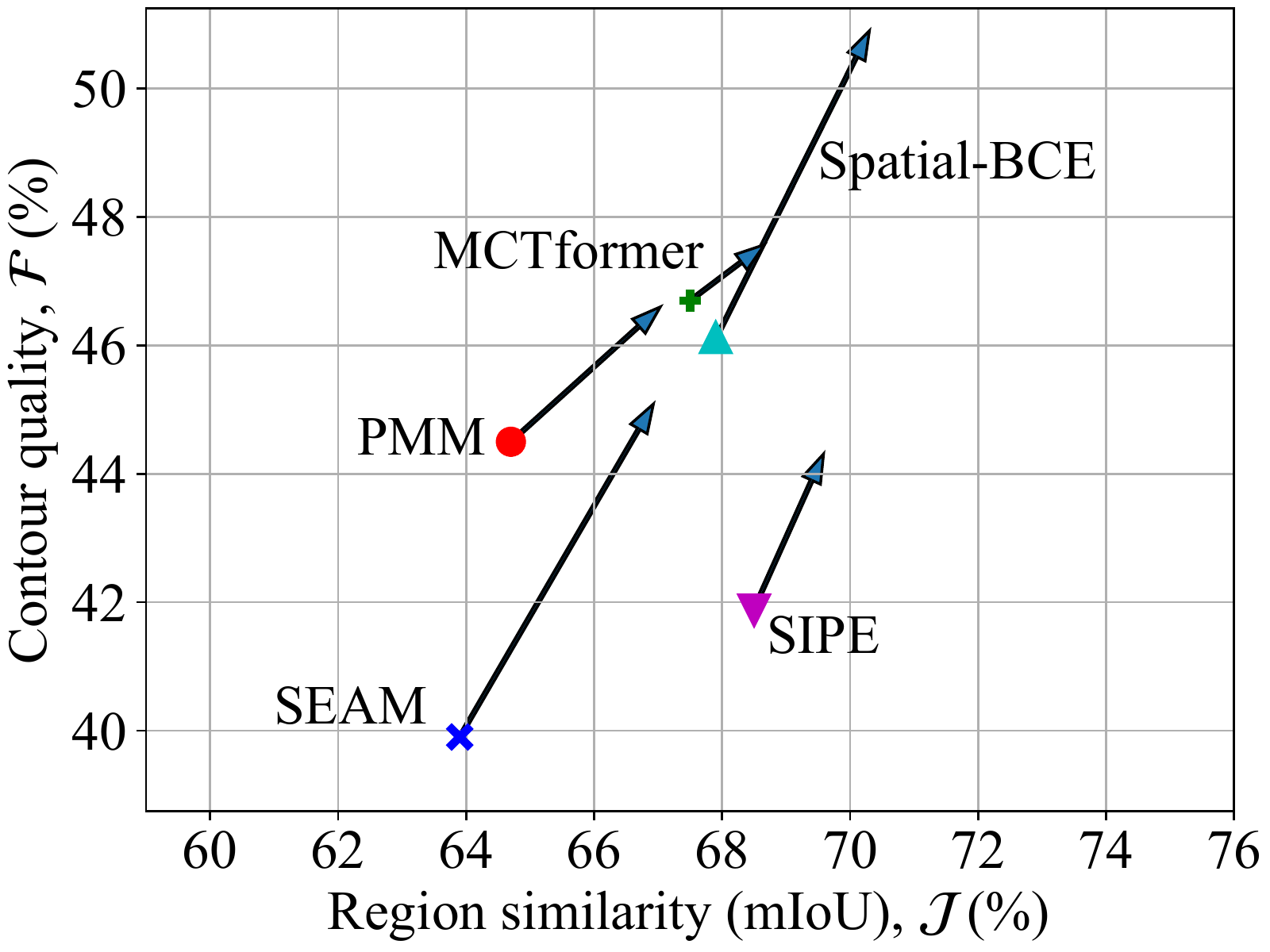}
    \vspace{-6pt}
    \caption{Region similarity vs.\ contour quality. The arrows represent improvements from our proposed approach, which boosts the performance of five state-of-the-art methods. The results are five-run averages, which required reimplementation of all methods.}
    \label{fig_j_vs_f}
    \vspace{-8pt}
\end{figure}

\begin{figure*}[!t]
    \centering
    \setlength{\tabcolsep}{0pt}
    \begin{tabular}{ccccc}
        \includegraphics[width=.2\linewidth]{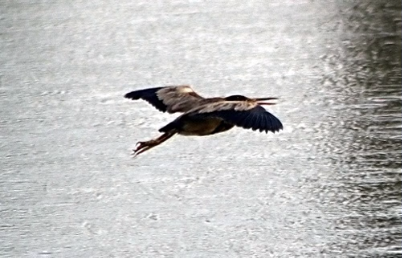} &
        \includegraphics[width=.2\linewidth]{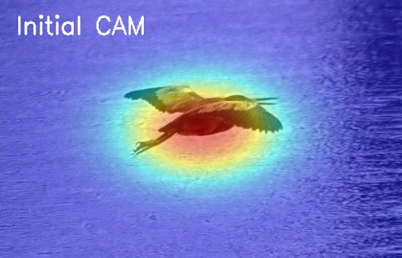} &
        \includegraphics[width=.2\linewidth]{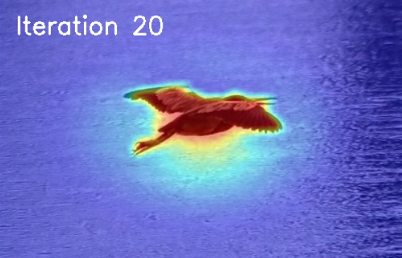} &
        \includegraphics[width=.2\linewidth]{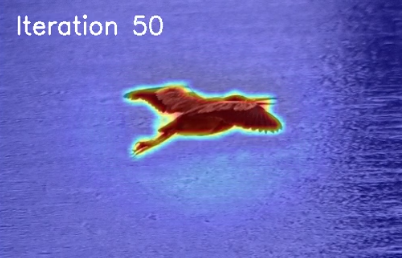}
    \end{tabular}
    \vspace{-10pt}
    \caption{Illustration of how our improved feature similarity loss improves an initial CAM.}
    \label{fig_fsl_demo}
    \vspace{-8pt}
\end{figure*}

\section{Introduction}

Recent years have witnessed significant advancements in deep learning methods, impacting many computer vision tasks, including semantic segmentation. Automatic image segmentation has proven useful in many applications, including autonomous driving \cite{cordts2016cityscapes}, video surveillance \cite{gruosso2021human}, and medical image analysis \cite{minaee2021image}. Although remarkable results have been achieved by fully-supervised segmentation frameworks, the utilization of large datasets with pixel-wise annotated images necessitates a significant manual labeling effort, which increases with the dataset size. To address this challenge, weakly-supervised semantic segmentation (WSSS) aims to alleviate the labelling effort by other alternatives, such as image-level labels \cite{wang2020cvpr}, bounding boxes \cite{dai2015iccv}, scribbles \cite{lin2016scribblesup}, and points \cite{bearman2016s}. Out of all the available options, image-level classification labels are the most cost-effective and can be obtained from various sources, making them the most popular choice, and addressed in this work.

WSSS with image-level supervision aims to produce pseudo-masks for images from classification labels only. For this purpose, class activation maps (CAMs) \cite{zhou2016cvpr} are exploited in many methods. While this approach has been shown to have a remarkable ability for localization, it does not take the object shape into account. This leads to activation mostly around discriminative regions and overly smooth CAMs, without well-defined prediction contours. This is due to the nature of global average pooling, where all the pixels in the image are treated as the same category, which is problematic in background areas. In the particular case of SEAM \cite{wang2020cvpr}, two complementary techniques have been proposed to overcome this drawback \cite{jonnarth2022icassp}: (1) The importance sampling loss (ISL), which applies the classification loss to pixels sampled from a distribution based on the CAM; (2) the feature similarity loss (FSL), which aligns prediction contours with edges in the image. However, they are based on the multinomial posterior with softmax, and implicitly assume that classes are mutually exclusive, which we find is suboptimal. We aim to address this drawback.

The mutual exclusivity assumption for ISL and FSL is suboptimal for two reasons: (1) It is violated in practice, as the losses are applied to downsampled CAMs, where a single pixel represents a larger image region. Thus, two classes can share the space occupied by a single pixel, especially near class borders. (2) The techniques do not easily generalize beyond the SEAM baseline, as most methods treat the classes independently based on the sigmoid activation. Moreover, some methods do not explicitly model the background, which is necessary for ISL with softmax, as the foreground probability otherwise sums to 1 in every pixel.

To solve both issues at once, we model multiple binary problems, with independent binomial posteriors for each class. We find that, as no assumption is made on class co-occurrences, the performance is improved. \Cref{fig_fsl_demo} highlights our improved FSL, where foreground and background are fully decoupled for an initial coarse CAM. Additionally, as this approach aligns with the majority of WSSS methods, our proposed method generalizes seamlessly, and boosts the performance of all tested state-of-the-art baselines, as shown in \cref{fig_j_vs_f}. Compared to the approach in the short paper \cite{jonnarth2022icassp}, our contributions are three-fold:
\begin{itemize}[leftmargin=*, topsep=2pt]
    \setlength\itemsep{-3pt}
    \item We analyse the weaknesses of the multinomial-based approach in \cite{jonnarth2022icassp} that limit performance and generality, restricting the method to the SEAM baseline.
    \item We model the independent binomial posteriors as opposed to the multinomial posterior, to formulate an add-on method that generalizes to virtually any baseline.
    \item In extensive ablations and experiments, we verify the choices in our approach and show that it improves several state-of-the-art methods on the common mIoU and the complementary contour quality, as shown in \cref{fig_j_vs_f}.
\end{itemize}

\section{Related work}

\textbf{Types of weak supervision.} Full supervision of semantic segmentation requires pixel-wise annotations, which are difficult to acquire and can suffer from inaccuracies. Therefore, it is of interest to investigate weak supervision that requires less manual labeling. Bounding boxes are a natural substitute for pixel-wise labels, and are exploited in different works \cite{papandreou2015iccv, khoreva2017cvpr, dai2015iccv, lee2021bbam}. Inspired by interactive image segmentation, scribbles \cite{lin2016scribblesup, vernaza2017cvpr} and points \cite{bearman2016eccv} have been used to ease the annotation. Image classification labels \cite{zhou2016cvpr}, which are the easiest to acquire, are also the most common.

\textbf{Pseudo-label generation} from image classification labels constitutes a core part of WSSS. Zhou \etal \cite{zhou2016cvpr} introduced class activation maps (CAMs) by revisiting the global average pooling (GAP) layer in classification networks, and showed that they had a remarkable localization ability, despite being trained with only classification labels. However, CAMs usually only activate over sparse discriminative regions. Subsequent works aim to overcome this drawback. Kolesnikov \etal \cite{kolesnikov2016eccv} introduced a global weighted rank pooling to replace GAP, and a constrain-to-boundary loss to learn precise boundaries during training. Wei \etal \cite{wei2017cvpr} proposed an adversarial erasing approach to mining and expanding object regions continuously. Huang \etal \cite{huang2018cvpr} utilized the classical seeded region growing method to expand the discriminative regions to cover the entire objects. Lee \etal \cite{lee2019cvpr} proposed FickleNet using dropout to stochastically select hidden units, allowing a single network to generate localization maps for different object parts. Wang \etal \cite{wang2020cvpr} proposed a self-supervised equivariant attention mechanism (SEAM) by imposing consistency regularization in a siamese network. Closely related to our work, Jonnarth and Felsberg \cite{jonnarth2022icassp} performed importance sampling as a substitute for GAP, and introduced an auxiliary feature similarity loss as additional supervision for SEAM \cite{wang2020cvpr}. The main differences are: (1) We analyse the weaknesses that limit their performance and generality; (2) we model the binomial as opposed to the multinomial posterior; (3) we generalize ISL and FSL beyond SEAM, to virtually any baseline.

\textbf{Pseudo-label refinement} methods enhance the pseudo-masks based on the CAMs. Ahn \etal \cite{ahn2018cvpr} proposed AffinityNet that predicts class-agnostic pixel-level semantic affinity. They further explored inter-pixel relations (IRN) \cite{ahn2019cvpr}, where confident seed areas are identified, and propagated via random walk. Chen \etal \cite{chen2020weakly} explored object boundaries explicitly and used the explored boundaries to constrain the localization map propagation. Lee \etal \cite{lee2021anti} introduced an anti-adversarial technique to manipulate images for increasing its classification score. Krähenbühl and Koltun \cite{krahenbuhl2011neurips} proposed a refinement method based on conditional random fields (CRF), which is commonly used in WSSS.

\section{Preliminaries}

\subsection{The weakly-supervised segmentation task}

We consider the image-level weakly-supervised semantic segmentation (WSSS) task. As opposed to the fully supervised case, no pixel-wise annotations are used for training. Given a dataset of images and corresponding image-level labels, we aim to train a model that predicts the semantic class for every pixel in an image. Commonly, this is achieved by first training a multi-label classification network, from which pseudo-labels are generated, commonly based on CAMs \cite{wang2020cvpr}. Optionally, a pseudo-label refinement method is applied \cite{ahn2018cvpr}. Finally, a segmentation model is trained in the fully-supervised setting, where the pseudo-labels are used as ground truth. Our contributions focus on the first stage of generating high-fidelity pseudo-labels.

\subsection{Class activation maps}

Class activation maps (CAM) of different forms have become a core component in the majority of weakly supervised segmentation methods \cite{qichen2022cvpr, zhaozhengchen2022cvpr, wang2020cvpr}. Zhou \etal \cite{zhou2016cvpr} introduced the CAM concept, and showed that classification networks that use a global average pooling (GAP) layer posses the ability to localize objects without any supervision on their locations, such as bounding boxes or segmentation masks. In their network architecture, they perform GAP on the feature map $f_k(i, j)$ over the spatial coordinates $i$ and $j$ for each channel $k$, where $f_k(i, j)$ is the output of the final convolutional layer. They then apply a single fully connected layer with weights $w$, and thus, the \textit{image-level} score for class $c$ is
\begin{equation}
    U_c = \sum_k w_k^c \sum_{i,j} f_k (i, j) = \sum_{i,j} \sum_k w_k^c f_k (i, j).
\end{equation}
The weight $w_k^c$ can be seen as the importance of feature $k$ on class $c$. Thus, they formulate the CAM as
\begin{equation}
    M_c(i, j) = \sum_k w_k^c f_k(i, j),
\end{equation}
which is the weighted sum over the features, and directly indicates the importance at spatial grid position $(i, j)$ \cite{zhou2016cvpr}.

Since CAMs provide the ability to localize objects using only classification labels, they are a natural choice for weakly supervised segmentation. However, most WSSS methods modify the original CAM, by removing the fully connected layer, and explicitly represent the output of the final convolutional layer as the spatial class score $s_c(i, j)$, with the same number of channels, $C$, as there are classes \cite{qichen2022cvpr, wang2020cvpr, xu2022cvpr}. In this case, the CAM is equivalent to $s_c$. GAP is then applied to $s_c$ to get the image-level class score
\begin{equation}
    \label{eq_gap}
    S_c = \sum_{i,j} s_c (i, j).
\end{equation}
Furthermore, in WSSS, the CAMs are usually normalized to the range $[0, 1]$ for pseudo-label generation. A common method involves ReLU and max normalization
\cite{qichen2022cvpr, wang2020cvpr, yoon2022eccv}
\begin{equation}
    \label{eq_relu_maxnorm}
    r_c(i, j) = \frac{ \mathrm{ReLU} (s_c(i, j))}{\max_{m,n} \mathrm{ReLU} (s_c(m, n))},
\end{equation}
which we refer to as max-normalized CAMs.

A third option for creating CAMs is to model the pixel-wise multinomial posterior, prior to global pooling \cite{jonnarth2022icassp}. The normalized CAMs are computed using softmax
\begin{equation}
    \label{eq_cam_softmax}
    a_c(i, j) = \frac{e^{s_c(i, j)}}{\sum_{k=1}^C e^{s_k(i, j)}} \cong \mathrm{Pr}(z_c(i, j) = 1 | x),
\end{equation}
given image $x$, where $C$ is the number of classes, $z_c(i, j)$ is the true class label, which is 1 if class $c$ occupies position $(i, j)$, or 0 otherwise. This is useful in the fully-supervised case, as the posterior can be directly supervised using ground-truth masks in full image resolution.

\subsection{Importance sampling loss}

Recently, importance sampling has been proposed as a substitute for global pooling during weakly supervised training \cite{jonnarth2022icassp}. It aims to solve the problem of global max pooling (GMP), which mainly targets the most discriminative region of an object. The authors in \cite{jonnarth2022icassp} model the multinomial posterior as in \eqref{eq_cam_softmax}, and define a probability mass function over pixel coordinates $(i, j)$;
\begin{equation}
    \label{eq_importance_sampling_pmf}
    p_c(I, J | x ) = \mathrm{Pr} (I=i, J=j | x, c) = a_c(i, j) / Z(a_c),
\end{equation}
where $Z(a_c) = \sum_{m=1}^W \sum_{n=1}^H a_c(m, n)$ is a normalizing constant for an image of width $W$ and height $H$. A pixel is then drawn from this distribution for each class, and the image-level prediction $\tilde{y}$ is computed as
\begin{equation}
    \label{eq_importance_sampling}
    \tilde{y}_c = a_c(\hat{\imath}, \hat{\jmath}), \quad (\hat{\imath}, \hat{\jmath}) \sim p_c(I, J | x ).
\end{equation}

Finally, the importance sampling loss (ISL) is the binary cross-entropy loss of the sampled prediction
\begin{equation}
    \label{eq_isl}
    \mathcal{L}_\mathrm{is} (y, \tilde{y}) = -\frac{1}{C} \sum_{c=1}^C y_c \log \tilde{y}_c + (1 - y_c) \log ( 1 - \tilde{y}_c),
\end{equation}
where $y$ is a vector of the image-level classification labels, and $y_c$ is the label for class $c$, which is 1 if class $c$ is present anywhere in the image, and 0 otherwise.

\subsection{Feature similarity loss}

\begin{figure}[t]
    \centering
    \includegraphics[width=0.79\columnwidth]{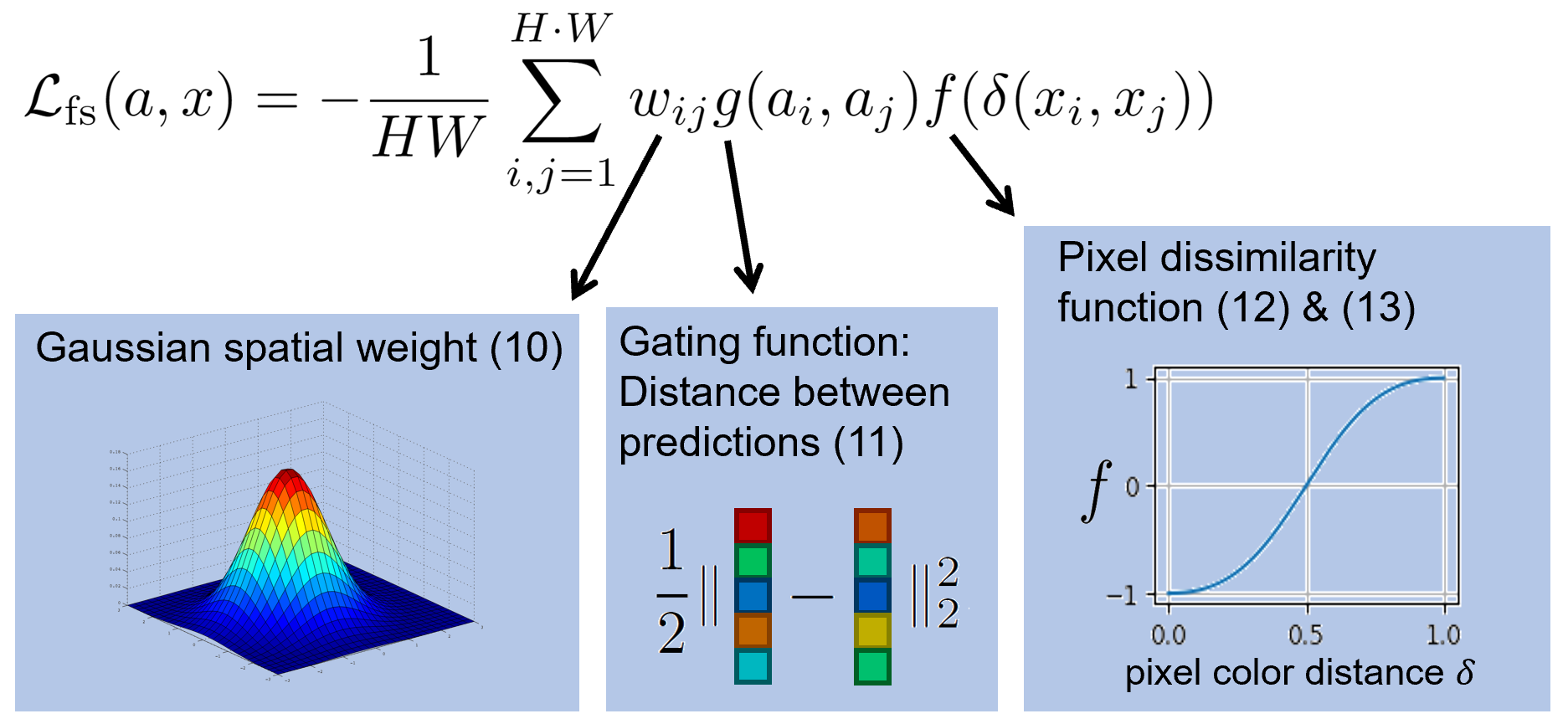}
    \vspace{-8pt}
    \caption{Illustration of the feature similarity loss (FSL).}
    \label{fig_fsl_illustration}
    \vspace{-6pt}
\end{figure}

Empirically, objects are in most cases separated from one another, or the background, by a color edge. Jonnarth and Felsberg \cite{jonnarth2022icassp} encapsulate this prior knowledge into the self-supervised feature similarity loss function
\begin{equation}
    \label{eq_fsl}
    \mathcal{L}_\mathrm{fs} (a, x) = -\frac{1}{HW} \sum_{i,j=1}^{H \cdot W} w_{ij} g(a_i, a_j) f(\delta(x_i, x_j)),
\end{equation}
which aligns the prediction contours with edges in the image. It is a function of the posterior $a \in [0, 1]^{H \times W \times C}$ in \eqref{eq_cam_softmax}, and the RGB image $x \in [0, 1]^{H \times W \times 3}$. Note that the spatial indexing differs from above. Similar to \cite{jonnarth2022icassp}, a single index is used for both of the spatial coordinates. The loss term sums over all \textit{pixel pairs}, and consists of three components. First, a Gaussian spatial weight is computed
\begin{equation}
    \label{eq_fsl_w}
    w_{ij} = \frac{1}{2 \pi \sigma^2} \exp \left( -\frac{|| p_i - p_j ||_2^2}{2 \sigma^2} \right),
\end{equation}
where $p_i$ is the two-dimensional position vector for pixel $i$. This assigns a higher importance to nearby pixel pairs. Very distant pairs are disregarded, as little to no conclusion can be drawn about their semantic affinity, solely based on color information. The second component is a gating function
\begin{equation}
    \label{eq_fsl_g}
    g(a_i, a_j) = \frac{1}{2} || a_i - a_j ||_2^2,
\end{equation}
which is the distance between the posteriors of pixels $i$ and $j$. The final component is a pixel dissimilarity function
\begin{align}
    \label{eq_fsl_f}
    & f(\delta) = \tanh \left( \mu + \log \left( \frac{\delta}{1 - \delta} \right) \right), \\
    \label{eq_fsl_delta}
    & \delta(x_i, x_j) = \frac{|| x_i - x_j ||_1}{3},
\end{align}
which maps two RGB pixel values to a dissimilarity score in $[-1, 1]$. See \cref{fig_fsl_illustration} for an illustration of FSL.

The gradient is only propagated through $g$, as $a$ is the network output. The weight $w_{ij}$ controls the magnitude of the loss, while $f$ controls the sign. A pair of similar pixels induces a positive loss, which is minimized by equal predictions. The opposite is true for dissimilar pixels. However, if the predictions are equal to begin with, then $\nabla \mathcal{L}_\mathrm{fs} = 0$, allowing consistent predictions over heterogeneous image regions. The two parameters $\sigma$ and $\mu$ are learned \cite{jonnarth2022icassp}.

\section{Modeling the binomial posterior}
\label{sec_likelihood_for_isl_fsl}

In this section, we first explain the problem of modeling the multinomial posterior for ISL and FSL, and proceed to present our binomial-based approach.

\subsection{The limits of multinomial-based ISL and FSL}
\label{sec_multinomial_limits}

The multinomial posterior in \eqref{eq_cam_softmax}, as used in the original ISL and FSL \cite{jonnarth2022icassp}, assumes that classes are mutually exclusive, \ie that only a single class can be present in each pixel. This assumption is reasonably sound in sufficiently high image resolution, where a single pixel is small enough to only house a single class, for all intents and purposes. Although, it may still be violated, \eg at class borders where the pixel-grid does not align with the object contours, or for distant objects where the size of an object is only a few pixels. However, the assumption becomes entirely void in downsampled CAMs, where the losses are applied, due to the fact that they are usually computed in a much lower resolution compared to the original input images, as a result of a fixed input size and downsampling layers in the network. This means that the assumption is not made on the pixel-level, but in fact, within a small region defined by the final downsampling scale factor. In summary, the mutual exclusivity assumption does not hold in practice. This leads to a loss in performance, as can be seen in \cref{table_ablation_likelihood_vs_posterior}.

Moreover, ISL and FSL cannot easily be applied to stronger baselines than SEAM \cite{wang2020cvpr}, since most methods compute independent class scores \cite{li2021iccv,su2021iccv}, and use GAP according to \eqref{eq_gap}. We found in our initial experiments that enforcing a multinomial posterior estimation with \eqref{eq_cam_softmax} resulted in performance degradation for other baselines. A likely explanation is that, as \eqref{eq_cam_softmax} changes the characteristics of the learned CAMs \cite{jonnarth2022icassp}, optimal hyperparameter values are affected. In fact, Jonnarth and Felsberg \cite{jonnarth2022icassp} found it necessary to heavily modify the background parameter $\alpha$ for AffinityNet \cite{ahn2018cvpr} label generation, since the CAM characteristics had changed. Similarly, hyperparameter values inherent to other baselines may also need to be adapted.

We argue that ISL and FSL would benefit from relaxing the mutual exclusivity assumption. Instead of looking at the multinomial, we consider the $C$-fold binary problem, which constitutes modeling $C$ independent binomial posteriors. We aim to adapt ISL and FSL based on this new formulation, so that they can be seamlessly implemented as add-on methods to any previous or future method, without any major modifications to the underlying baselines. Note that, in the case of unbalanced training data, the multiple binomial predictors are assumed to be calibrated, \eg by means of logit-adjustment \cite{menon2020adjustment,aimar2023balanced}.

\subsection{Binomial-based importance sampling loss}
\label{sec_improving_isl}

A natural approach is to interpret the class scores, $s_c$, as log-odds, or logits, and model the pixel-wise binomial posterior using the sigmoid function, which is given by
\begin{equation}
    \label{eq_cam_sigmoid}
    b_c(i, j) = \frac{e^{s_c(i, j)}}{1 + e^{s_c(i, j)}} \cong \mathrm{Pr}(z_c(i, j) = 1 | x).
\end{equation}
The sampling distribution is attained in a similar manner as for the multinomial posterior case, namely
\begin{equation}
    p_c(I, J | x ) = b_c(i, j) / Z(b_c).
\end{equation}
We sample the binomial $\tilde{b}_c = b_c(\hat{\imath}, \hat{\jmath})$, where $(\hat{\imath}, \hat{\jmath}) \sim p_c$, and the final loss is thus computed as $\mathcal{L}_{\mathrm{is}} (y, \tilde{b})$ using \eqref{eq_isl}.

Moreover, importance sampling has been shown to be superior to GMP \cite{jonnarth2022icassp}. However, Zhou \etal \cite{zhou2016cvpr} found that while GMP achieves similar classification performance as GAP, GAP outperforms GMP for localization. Thus, we define the final classification loss as a convex combination of the importance sampling loss and the binary cross-entropy loss, $\mathcal{L}_\mathrm{ce}$, based on GAP, similarly to how the original authors used GMP. Our combined classification loss is
\begin{align}
    \label{eq_cls_loss}
    & \mathcal{L}_{\mathrm{cls}} = (1 - \lambda) \mathcal{L}_{\mathrm{ce}} + \lambda \mathcal{L}_{\mathrm{is}}, \\
    \label{eq_ce_loss}
    & \mathcal{L}_\mathrm{ce} = -\frac{1}{C} \sum_{c=1}^C y_c \log B_c + (1 - y_c) \log (1 - B_c),
\end{align}
where $B_c = \exp (S_c) / (1 + \exp (S_c))$ is the \textit{image-level} GAP-based posterior estimate based on $S_c$ in \eqref{eq_gap}.

\begin{figure}[t]
    \centering
    \includegraphics[width=0.62\columnwidth]{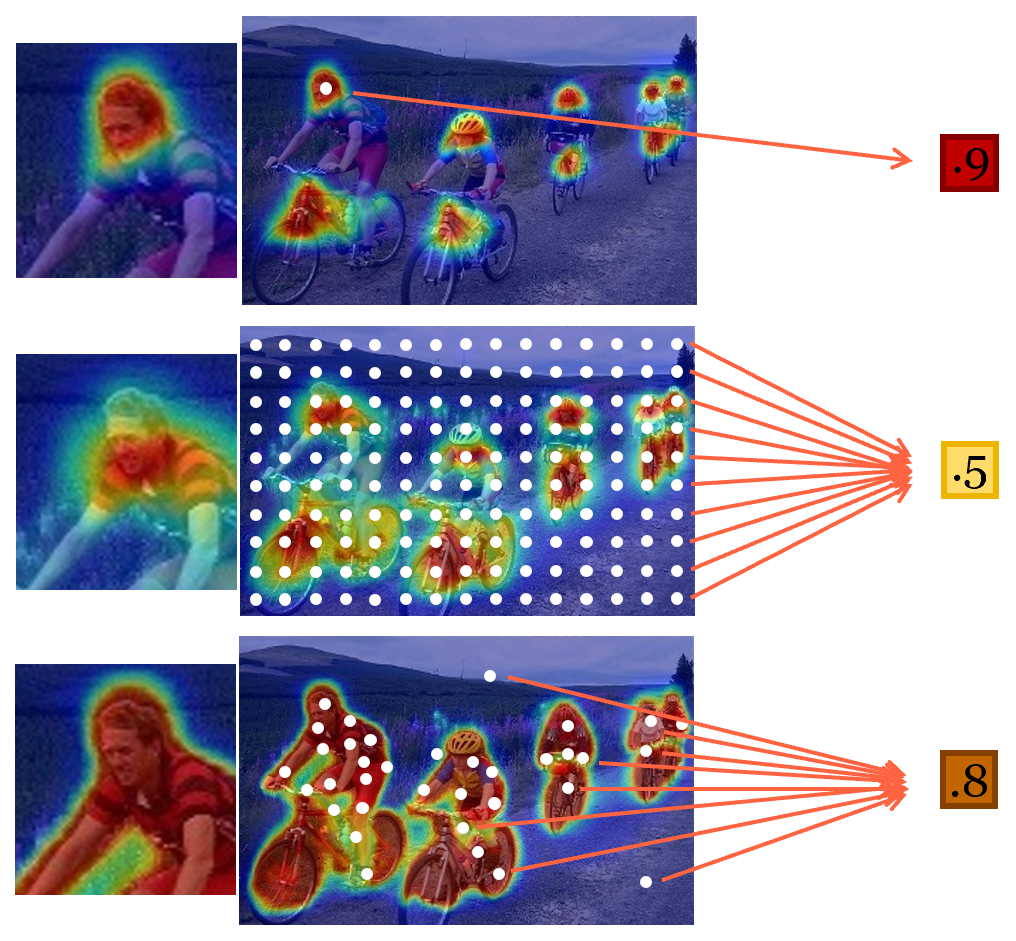}
    \vspace{-8pt}
    \caption{Illustration of global pooling methods. Top to bottom; max pooling, average pooling, multi-sample importance sampling.}
    \label{fig_isl_illustration}
\end{figure}

Finally, to improve stability, we propose to sample multiple pixels per class, and average their loss contributions. We write the multi-sample importance sampling loss as
\begin{equation}
    \mathcal{L}_\mathrm{is}^N = \frac{1}{N} \sum_{n=1}^N \mathcal{L}_\mathrm{is} (y, \tilde{b}^{(n)}),
\end{equation}
where $N$ is the number of samples drawn per class, and $\tilde{b}^{(n)}$ for $n = 1, ..., N$ are prediction vectors of length $C$, containing independently drawn samples. \Cref{fig_isl_illustration} illustrates the difference between GMP, GAP, and multi-sample ISL.

\subsection{Binomial-based feature similarity loss}
\label{sec_likelihood_fsl}

The mutual exclusivity assumption made when modeling the multinomial posterior influences the feature similarity loss. For dissimilar pixels, not only are the predictions pushed apart between \textit{pixels}, but also between \textit{classes}. Thus, the predicted class distribution will tend towards a one-hot vector \cite{jonnarth2022icassp}. Based on the argument in \cref{sec_multinomial_limits}, \ie that multiple classes can occupy the same pixel, we argue that classes should be handled independently when reasoning about their contours. In fact, we believe that this reasoning is even more important at class borders, where two classes are likely to occupy the same pixel, especially in downsampled CAMs where FSL is applied.

Thus, we adapt FSL for binomial modeling according to \eqref{eq_cam_sigmoid}, where the classes are considered independently, in the absence of softmax. We consider three different inputs to the gating function $g$ in place of the multinomial posterior $a$: The raw scores $s$, the binomial $b$ in \eqref{eq_cam_sigmoid}, and the max-normalized CAM $r$ in \eqref{eq_relu_maxnorm}. A closer look at the gradients of $g$ in \eqref{eq_fsl_g} reveals that, when using the scores $s$ directly,
\begin{equation}
    \frac{\partial g(s_i, s_j)}{\partial s_{i,c}} = s_{i,c} - s_{j,c}.
\end{equation}
The gradient is proportional to the score difference. For dissimilar pixels, $g$ is maximized, which leads to divergence, as the gradients increase indefinitely. The gradient is bounded for the binomial posterior $b$, we get
\begin{align}
    \left| \frac{\partial g(b_i, b_j)}{\partial s_{i,c}} \right| & = \left| (b_{i,c} - b_{j,c}) \frac{e^{s_{i,c}}}{(1 + e^{s_{i,c}})^2} \right| \\
    & \leq e^{s_{i,c}} / (1 + e^{s_{i,c}})^2 \leq 1 / e^{s_{i,c}}.
\end{align}
The same is true for $r$. Assuming $s_{i,c} \geq 0$, we get
\begin{align}
    \left| \frac{\partial g(r_i, r_j)}{\partial s_{i,c}} \right| & = \left| (r_{i,c} - r_{j,c}) \frac{1}{\mathrm{max}_m (s_{m,c})} \right| \\
    & \leq 1 / \mathrm{max}_m (s_{m,c}).
\end{align}
Since the gradients are bounded for $b$ and $r$, the scores do not diverge out of control whenever the prediction difference is large. Note that for $r$, if $s_{i,c} < s_{j,c}$ the score $s_{i,c}$ will decrease towards 0, and if $s_{i,c} < 0$ the gradient is 0. This means that if the model predicts that a class is absent, \ie the score is negative, the contours are disregarded. There is no analogous lower score bound for $b$. Therefore, we use $r$ for our experiments. This choice is verified in an ablation.

Furthermore, since the classes are considered independently, it no longer makes sense to compute the loss over classes that are not present. Their scores should be minimized over the entire image, which is handled by the classification loss. Enforcing prediction contours may only hurt performance, so we mask out these classes and only apply FSL on classes that are present in the image.

Finally, we find suitable FSL parameters and keep them fixed for all experiments, in contrast to learning them in \cite{jonnarth2022icassp}, as this can lead to trivial solutions. For instance, the gradient with respect to $\mu$ is always negative, so $\mu$ will only increase during training. This might yield inconsistent results if the training time is changed, \eg for large datasets with more gradient steps. We optimize $\mathcal{L}_\mathrm{fs}$ alone over initial Gaussian CAMs. In the spirit of WSSS, we only use one image per class to reduce the required ground-truth masks. The resulting parameter values were $\mu=2.5$ and $\sigma=5$. For further details, see the supplementary material.

\section{Experiments}

\subsection{Implementation details}
\label{sec_implementation_details}

\textbf{Datasets.} We test our approach on two datasets; \textit{PASCAL Visual Object Classes} (VOC) \cite{everingham2010ijvc}, and the larger \textit{MS Common Objects in Context} (COCO) \cite{lin2014eccv}. VOC contains 1,464 training images, 1,449 validation images, 1,456 hold-out test images, and includes 20 foreground classes. Following the common practice, we include additional training data from SBD \cite{hariharan2011iccv}, increasing the number of training images to 10,582. For COCO, we use the 2014 train-validation split, which contains 82,783 training images, 40,504 validation images, and includes 80 classes. Note that, although pixel-level segmentation ground truth is available for the training and validation sets, they are not used for training.

\textbf{Evaluation metrics.} For evaluation, we use two complementary metrics as in \cite{jonnarth2022icassp}. (1) Region similarity, $\mathcal{J}$, which is commonly used in WSSS. It compares segmentation masks in terms of the Jaccard index \cite{jaccard1901svsn}, which is the mean intersection over union (mIoU). (2) We complement this by also evaluating the contour quality, $\mathcal{F}$, drawing inspiration from the well-established DAVIS benchmark for video object segmentation \cite{perazzi2016cvpr}. This is the F-score computed on the segmentation mask contours, which is efficiently approximated using morphological operations to accumulate bipartite matches of the boundaries between the predicted and ground-truth masks. Finally, we compute a combined metric, $\mathcal{J}\&\mathcal{F}$, which is simply an average of $\mathcal{J}$ and $\mathcal{F}$. Mean and standard deviation is reported over five runs whenever confidence intervals are given. Note that this required reimplementation of all methods, as confidence intervals are not commonly reported in the WSSS literature.

\textbf{Training details.} Since we apply ISL and FSL to a wide variety of state-of-the-art baselines, we mainly use the default settings and parameters stated in the respective papers or linked code repositories. Moreover, we use the implementations referenced to in the papers, as is. Any deviations from the default settings are listed in the subsequent paragraphs. Note that we did not manage to reproduce the reported results exactly in all cases. Potential reasons include different software and hardware configurations, and the use of different evaluation code. We believe that a likely reason is that WSSS typically involves multiple steps spread out over several code repositories, which complicates the reproduction of results. For a further discussion, see the supplementary material. Still, our experiments provide good comparisons as they show how ISL and FSL can boost the performance under the same conditions as the baselines, and as the reproduced results are close to the reported results in the literature, which is evident in \cref{table_cam}. We used four A100 40GB GPUs in our experiments.

\textbf{SEAM} \cite{wang2020cvpr}. We set the background parameter to 1 and 24 (originally 4 and 24) for AffinityNet, when using ISL. On COCO, FSL was scaled down by a factor of $0.2$.

\textbf{SIPE} \cite{qichen2022cvpr}. ISL and FSL were scaled down by a factor of $0.1$, since SIPE uses a higher learning rate by default. For FSL, we upsample the CAMs by a factor of 2 (from $32 \times 32$ to $64 \times 64$), to be closer to SEAM ($56 \times 56$). We chose an integer factor to avoid artifacts. Note that the ResNet-101 implementation for VOC referenced by SIPE \cite{qichen2022cvpr} uses COCO pretrained weights, which means that ground-truth segmentation masks were used, and thus, these reproduced results are not weakly supervised strictly speaking.

\textbf{PMM} \cite{li2021iccv}. We use ISL only during multi-scale training, and FSL during both multi-scale and multi-crop training.

\textbf{MCTformer} \cite{xu2022cvpr}. ISL and FSL were scaled down by a factor of $0.1$, since MCTformer uses the AdamW optimizer \cite{loshchilov2019iclr} instead of SGD. For FSL, we upsample the CAMs by a factor of 4 (from $14 \times 14$ to $56 \times 56$) to match SEAM.

\textbf{Spatial-BCE} \cite{wu2022adaptive}. Same as the original setting, the foreground and background pixel thresholds in IRN \cite{ahn2019cvpr} were set to 0.60 and 0.25 respectively. The semantic segmentation threshold was set to 0.30 with ISL, and 0.40 with FSL.

\subsection{ISL and FSL improve pseudo-label fidelity}

In \cref{table_cam} we evaluate pseudo-label fidelity by comparing region similarity of pseudo-labels generated from CAMs. We find that our losses have a significant effect on the pseudo-labels, resulting in an average relative performance increase of $+5\%$. Note that we even managed to boost the performance of a transformer-based method, MCTformer \cite{xu2022cvpr}, despite its low CAM resolution of $14 \times 14$, and architectural differences to the CNN-based methods. Not only that, but our losses seem to stabilize its training, reducing the standard deviation over five runs from $2.0$ to $0.4$.
While our losses increase the computational effort by $30\%$ for the overall training pipeline, the inference time is unaffected.
We further evaluate the impact of our improved pseudo-labels on final semantic segmentation performance in the weakly supervised setting.

\begin{table}[t]
    \small
    \renewcommand{\arraystretch}{0.9}
    \setlength{\tabcolsep}{4pt}
    \begin{center}
    \caption{Region similarity for CAM pseudo-labels without CRF on VOC. Max and mean over five runs are shown together with the reported results from the respective papers (\textit{``Ref.''} column).}
    \vspace{-6pt}
    \label{table_cam}
    \begin{tabular}{llccccc}
    \toprule
    & & & \multicolumn{4}{c}{Reproduced} \\
    \cmidrule(l{\tabcolsep}r{\tabcolsep}){4-7}
    & & & \multicolumn{2}{c}{no ISL/FSL} & \multicolumn{2}{c}{with ISL/FSL} \\
    \cmidrule(l{\tabcolsep}r{\tabcolsep}){4-5}
    \cmidrule(l{\tabcolsep}r{\tabcolsep}){6-7}
    Method & Set & Ref. & max & mean{\tiny $\pm$std} & max & mean{\tiny $\pm$std} \\
    \midrule
    SEAM \cite{wang2020cvpr} & \textit{train} & 55.4 & 55.6 & 55.0{\tiny $\pm$0.4} & 61.0 & 60.6{\tiny $\pm$0.3} \\
    SIPE \cite{qichen2022cvpr} & \textit{train} & 58.6 & 58.7 & 58.5{\tiny $\pm$0.2} & 60.4 & 60.1{\tiny $\pm$0.2} \\
    PMM \cite{li2021iccv} & \textit{val} & 56.2 & 55.4 & 55.1{\tiny $\pm$0.2} & 59.3 & 58.9{\tiny $\pm$0.3} \\
    \scriptsize MCTformer \small \cite{xu2022cvpr} & \textit{train} & 61.7 & 61.8 & 59.4{\tiny $\pm$2.0} & 61.6 & 61.2{\tiny $\pm$0.4} \\
    \scriptsize Spatial-BCE \small \cite{wu2022adaptive} & \textit{train} & 68.1 & 67.2 & 67.0{\tiny $\pm$0.2} & 68.7 & 68.6{\tiny $\pm$0.1} \\
    \bottomrule
    \end{tabular}
    \end{center}
    \vspace{-10pt}
\end{table}

\subsection{Segmentation comparison on VOC}

\cref{fig_j_vs_f} and \cref{table_seam} show our reproduced final segmentation results on a wide variety of state-of-the-art baselines, including three different backbones. Our improved pseudo-labels translate into more accurate segmentation predictions, consistently outperforming the baselines. On average, $\mathcal{J}$, $\mathcal{F}$, and $\mathcal{J}\&\mathcal{F}$ are increased by +2.2\%, +5.9\%, and +3.7\%, respectively, on the validation set. On the hold-out test set, we evaluate the best model out of five runs for each method, based on validation set performance. With ISL and FSL, the test set performance is improved by +1.6\% on average. Note that, the original MCTformer \cite{xu2022cvpr} had a large standard deviation, and a larger \textit{maximum} performance over five runs compared to when ISL and FSL were used. Thus, the test set performance was higher for the original method, although the \textit{average} performance was higher with ISL and FSL. This is consistent with \cref{table_cam}. This observation is important for practitioners who simply apply an off-the-shelf training method. A lower variance is beneficial, as the performance deviation is smaller, and the resulting model is more likely to perform as expected. Ideally, we would prefer to compute an average test set performance. However, this is prohibited by the limited number of test set submissions on VOC. Additional results are included in the supplementary material, and qualitative results are shown in \cref{fig_qualitative}.

\subsection{Segmentation comparison on COCO}

In \cref{table_sota_coco}, we evaluate our losses on the COCO dataset, which contains roughly 10 times more images than VOC. For SEAM \cite{wang2020cvpr}, the segmentation results are improved in this setting as well, demonstrating that our losses are well-suited for large-scale datasets. When training the classification network for 8 epochs, we boost $\mathcal{J}$ and $\mathcal{F}$ by +2.5\%. Our method further benefits from longer training, increasing $\mathcal{J}$ and $\mathcal{F}$ by an additional +2.8\% and +0.7\% respectively, while SEAM is less affected by the longer training.

\subsection{Ablation study}

For further analysis, we perform an ablation study investigating the effect of using the binomial versus multinomial, the number of sampled pixels, the loss weight $\lambda$, and the choice of gating function, with SEAM \cite{wang2020cvpr} as the baseline.

\begin{table}[t]
    \small
    \renewcommand{\arraystretch}{0.9}
    \setlength{\tabcolsep}{4pt}
    \begin{center}
        \caption{Final segmentation performance on VOC, in terms of region similarity ($\mathcal{J}$), contour quality ($\mathcal{F}$), and combined ($\mathcal{J}\&\mathcal{F}$).}
        \vspace{-6pt}
        \label{table_seam}
        \begin{tabular}{llccccc}
            \toprule
            & & \multicolumn{3}{c}{\textit{validation}} & \multicolumn{1}{c}{\textit{test}} \\
            
            \cmidrule(l{\tabcolsep}r{\tabcolsep}){3-5}
            \cmidrule(l{\tabcolsep}r{\tabcolsep}){6-6}
            Method & Backb. & $\mathcal{J}$ & $\mathcal{F}$ & $\mathcal{J}\&\mathcal{F}$ & $\mathcal{J}$ \\
            \midrule
            SEAM \cite{wang2020cvpr} & Res38 & 63.9{\tiny $\pm$0.5} & 39.9{\tiny $\pm$0.2} & 51.9{\tiny $\pm$0.3} & 65.4 \\
            \small{\textbf{+ISL/FSL (ours)}} & Res38 & \textbf{66.7}{\tiny $\pm$0.2} & \textbf{44.7}{\tiny $\pm$0.1} & \textbf{55.7}{\tiny $\pm$0.1} & \textbf{67.6} \\
            \midrule
            SIPE \cite{qichen2022cvpr} & Res38 & 68.0{\tiny $\pm$0.2} & 45.1{\tiny $\pm$0.2} & 56.6{\tiny $\pm$0.1} & 68.9 \\
            \small{\textbf{+ISL/FSL (ours)}} & Res38 & \textbf{68.3}{\tiny $\pm$0.2} & \textbf{46.8}{\tiny $\pm$0.2} & \textbf{57.6}{\tiny $\pm$0.1} & \textbf{69.4} \\
            \midrule
            SIPE \cite{qichen2022cvpr} & Res101 & 68.5{\tiny $\pm$0.2} & 41.9{\tiny $\pm$0.5} & 55.2{\tiny $\pm$0.3} & 69.4 \\
            \small{\textbf{+ISL/FSL (ours)}} & Res101 & \textbf{69.4}{\tiny $\pm$0.2} & \textbf{43.9}{\tiny $\pm$0.2} & \textbf{56.7}{\tiny $\pm$0.1} & \textbf{70.3} \\
            \midrule
            PMM \cite{li2021iccv} & Res38 & 64.7{\tiny $\pm$0.5} & 44.5{\tiny $\pm$0.5} & 54.6{\tiny $\pm$0.4} & 65.7 \\
            \small{\textbf{+ISL/FSL (ours)}} & Res38 & \textbf{66.7}{\tiny $\pm$0.7} & \textbf{46.3}{\tiny $\pm$0.6} & \textbf{56.5}{\tiny $\pm$0.6} & \textbf{67.0} \\
            \midrule
            \small{MCTformer} \cite{xu2022cvpr} & DeiT-S & 67.5{\tiny $\pm$1.7} & 46.7{\tiny $\pm$0.7} & 57.1{\tiny $\pm$1.2} & \textbf{70.6} \\
            \small{\textbf{+ISL/FSL (ours)}} & DeiT-S & \textbf{68.3}{\tiny $\pm$0.7} & \textbf{47.3}{\tiny $\pm$0.3} & \textbf{57.8}{\tiny $\pm$0.5} & 70.0 \\
            \midrule
            \small{Spatial-BCE} \cite{wu2022adaptive} & Res38 & 68.1{\tiny $\pm$0.1} & 45.4{\tiny $\pm$0.1} & 56.7{\tiny $\pm$0.1} & 68.4 \\
            \small{\textbf{+ISL/FSL (ours)}} & Res38 & \textbf{69.3}{\tiny $\pm$0.2} & \textbf{48.2}{\tiny $\pm$0.1} & \textbf{58.8}{\tiny $\pm$0.1} & \textbf{69.4} \\
            \midrule
            \small{Spatial-BCE} \cite{wu2022adaptive} & Res101 & 67.9{\tiny $\pm$0.2} & 46.1{\tiny $\pm$0.1} & 57.0{\tiny $\pm$0.1} & 68.4 \\
            \small{\textbf{+ISL/FSL (ours)}} & Res101 & \textbf{70.1}{\tiny $\pm$0.2} & \textbf{50.5}{\tiny $\pm$0.2} & \textbf{60.3}{\tiny $\pm$0.2} & \textbf{70.6} \\
            \bottomrule
        \end{tabular}
    \end{center}
    \vspace{-10pt}
\end{table}

\begin{table}[t]
    \small
    \renewcommand{\arraystretch}{0.9}
    \setlength{\tabcolsep}{4pt}
    \begin{center}
    \caption{Final segmentation performance on the COCO validation set, in terms of region similarity ($\mathcal{J}$), contour quality ($\mathcal{F}$), and combined ($\mathcal{J}\&\mathcal{F}$). The classification networks were trained for 8 epochs (three runs) and 16 epochs (single run with fixed seed).}
    \vspace{-6pt}
    \label{table_sota_coco}
    \begin{tabular}{lcccc}
    \toprule
    Method & Epochs & $\mathcal{J}$ & $\mathcal{F}$ & $\mathcal{J}\&\mathcal{F}$ \\
    \midrule
    SEAM~\cite{wang2020cvpr} & 8 & 35.9{\tiny $\pm$0.2} & 28.2{\tiny $\pm$0.3} & 32.1{\tiny $\pm$0.3} \\
    \textbf{+ISL/FSL (ours)} & 8 & \textbf{36.8}{\tiny $\pm$0.2} & \textbf{28.9}{\tiny $\pm$0.1} & \textbf{32.9}{\tiny $\pm$0.1} \\
    \midrule
    SEAM~\cite{wang2020cvpr} & 16 & 35.5 & 28.5 & 32.0 \\
    \textbf{+ISL/FSL (ours)} & 16 & \textbf{37.8} & \textbf{29.1} & \textbf{33.4} \\
    \bottomrule
    \end{tabular}
    \end{center}
    \vspace{-10pt}
\end{table}

\begin{table}[t]
    \small
    \renewcommand{\arraystretch}{0.9}
    \setlength{\tabcolsep}{4pt}
    \begin{center}
    \caption{Comparison between modeling the multinomial posterior in \eqref{eq_cam_softmax} versus the binomial posterior in \eqref{eq_cam_sigmoid} for ISL and FSL, with SEAM \cite{wang2020cvpr}. The region similarity ($\mathcal{J}$), contour quality ($\mathcal{F}$), and averaged ($\mathcal{J}$\&$\mathcal{F}$) are computed on the VOC validation set.}
    \vspace{-6pt}
    \label{table_ablation_likelihood_vs_posterior}
    \begin{tabular}{llccc}
    \toprule
    Loss & Model & $\mathcal{J}$ & $\mathcal{F}$ & $\mathcal{J}$\&$\mathcal{F}$ \\
    \midrule
    \multirow{2}{*}{ISL} & Multinomial \cite{jonnarth2022icassp} & 63.0{\tiny $\pm$0.5} & 41.0{\tiny $\pm$0.4} & 52.0{\tiny $\pm$0.4} \\
    & Binomial (proposed) & \textbf{64.3}{\tiny $\pm$0.8} & \textbf{42.2}{\tiny $\pm$0.6} & \textbf{53.3}{\tiny $\pm$0.7} \\
    \midrule
    \multirow{2}{*}{FSL} & Multinomial \cite{jonnarth2022icassp} & 50.3{\tiny $\pm$13} & 41.4{\tiny $\pm$8.5} & 45.9{\tiny $\pm$11} \\
    & Binomial (proposed) & \textbf{66.9}{\tiny $\pm$0.4} & \textbf{43.5}{\tiny $\pm$0.2} & \textbf{55.2}{\tiny $\pm$0.2} \\
    \bottomrule
    \end{tabular}
    \end{center}
    \vspace{-13pt}
\end{table}

\begin{figure*}[!t]
    \centering
    \renewcommand{\arraystretch}{0.3}
    \setlength{\tabcolsep}{0.5pt}
    \begin{tabular}{cccccccccccc}
        \includegraphics[width=.08\linewidth]{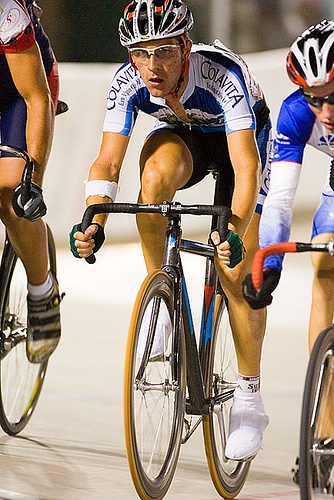} &
        \includegraphics[width=.08\linewidth]{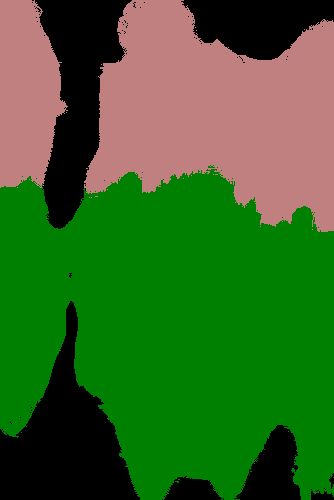} &
        \includegraphics[width=.08\linewidth]{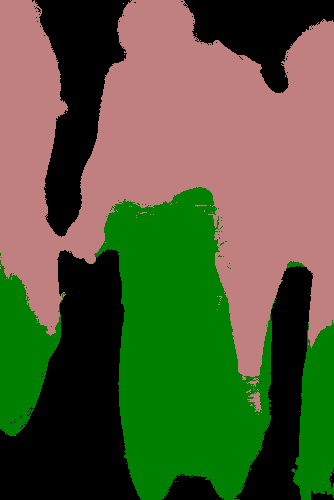} &
        \includegraphics[width=.08\linewidth]{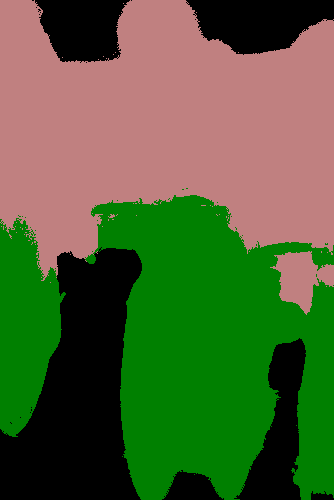} &
        \includegraphics[width=.08\linewidth]{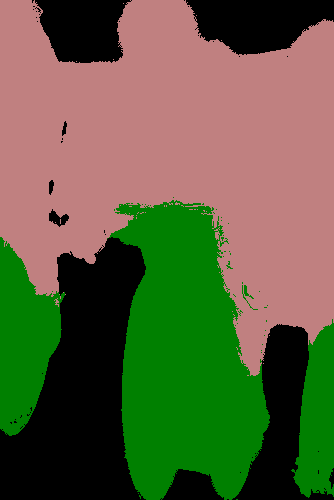} &
        \includegraphics[width=.08\linewidth]{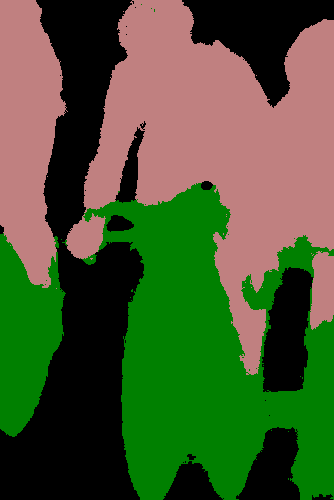} &
        \includegraphics[width=.08\linewidth]{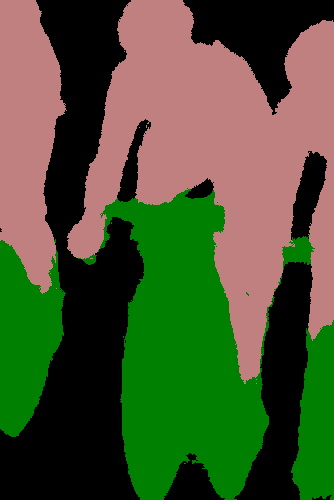} &
        \includegraphics[width=.08\linewidth]{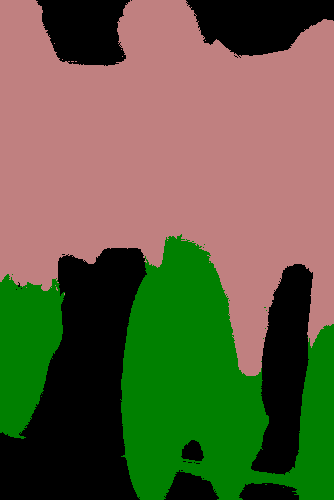} &
        \includegraphics[width=.08\linewidth]{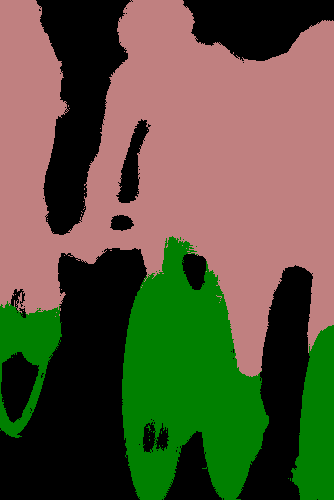} &
        \includegraphics[width=.08\linewidth]{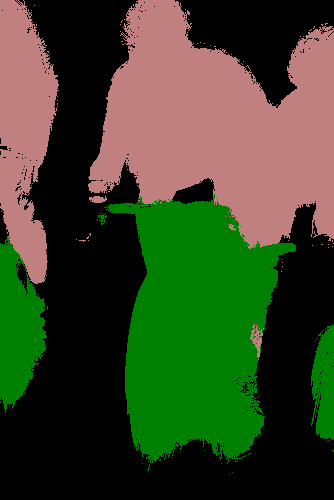} &
        \includegraphics[width=.08\linewidth]{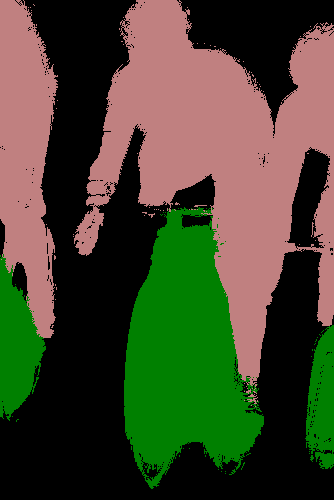} &
        \includegraphics[width=.08\linewidth]{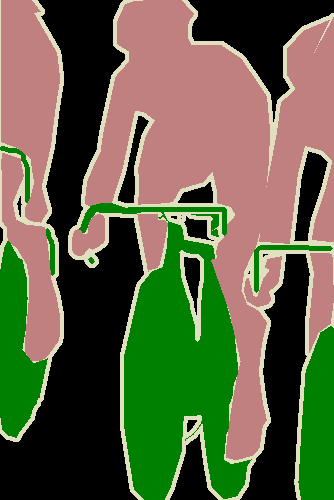} \\
        \includegraphics[width=.08\linewidth]{figures/quantitative_results/2007_000783_image.jpg} &
        \includegraphics[width=.08\linewidth]{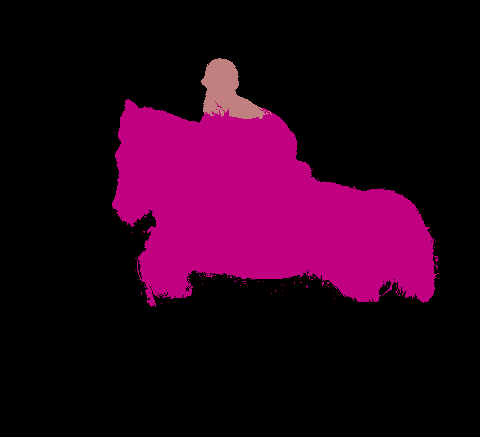} &
        \includegraphics[width=.08\linewidth]{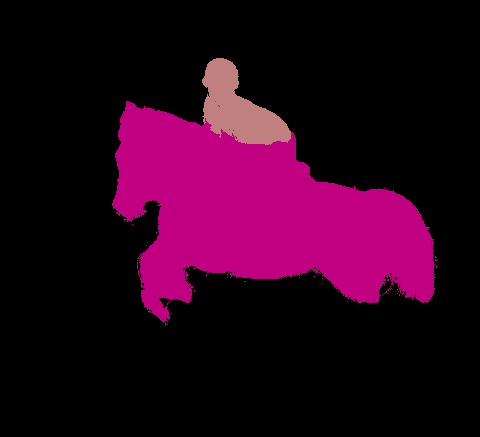} &
        \includegraphics[width=.08\linewidth]{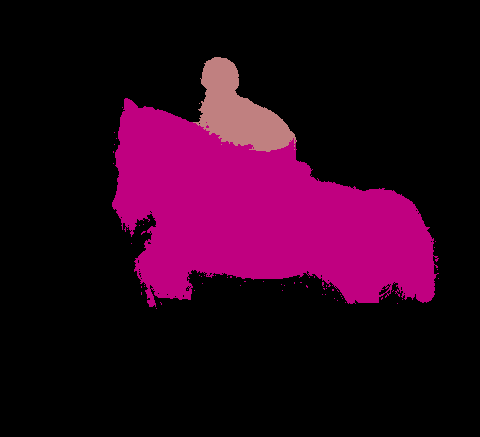} &
        \includegraphics[width=.08\linewidth]{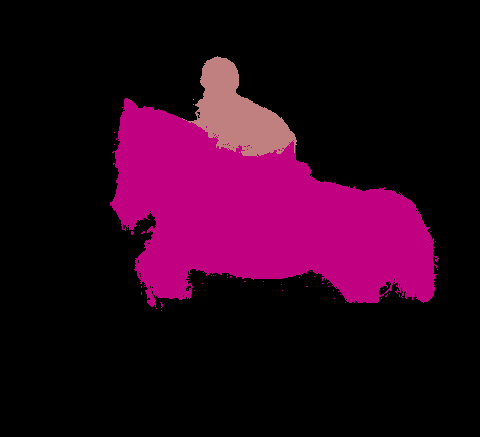} &
        \includegraphics[width=.08\linewidth]{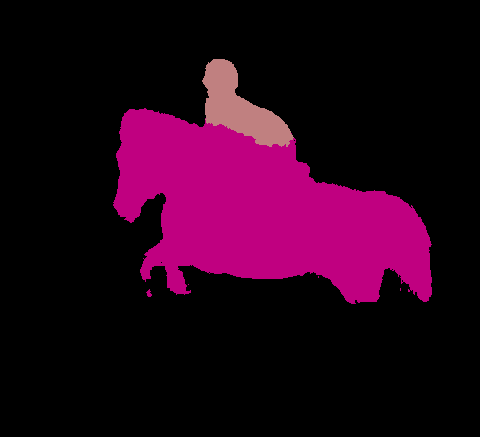} &
        \includegraphics[width=.08\linewidth]{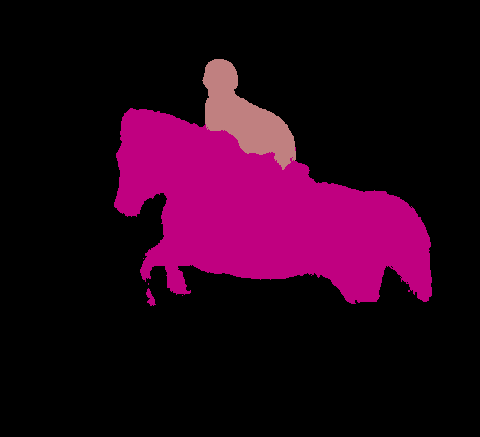} &
        \includegraphics[width=.08\linewidth]{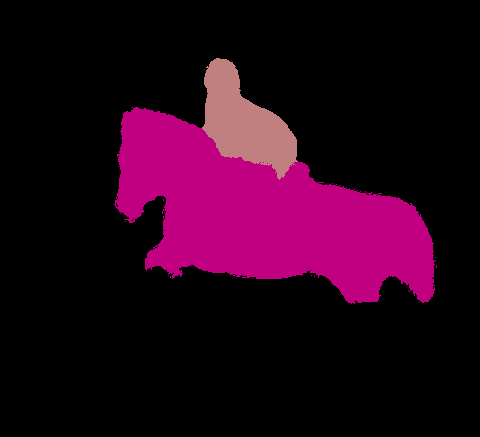} &
        \includegraphics[width=.08\linewidth]{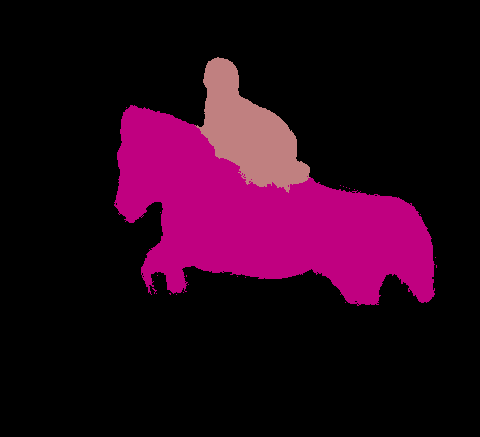} &
        \includegraphics[width=.08\linewidth]{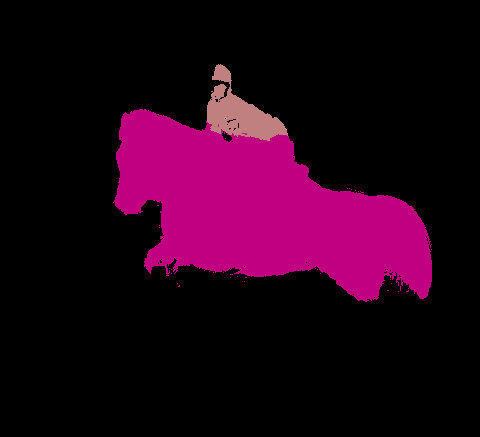} &
        \includegraphics[width=.08\linewidth]{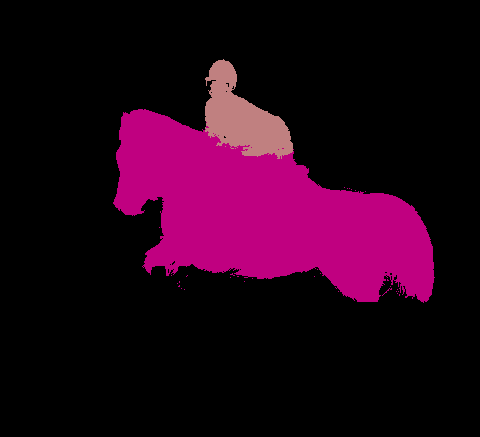} &
        \includegraphics[width=.08\linewidth]{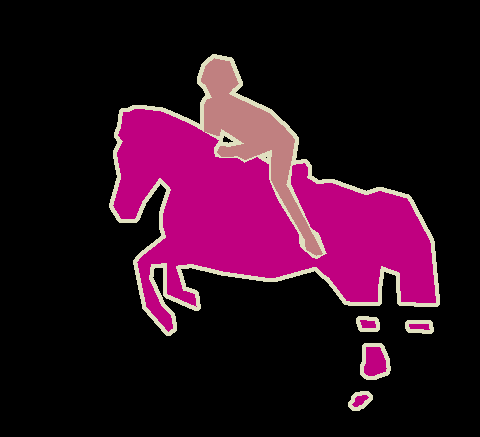} \\
        \includegraphics[width=.08\linewidth]{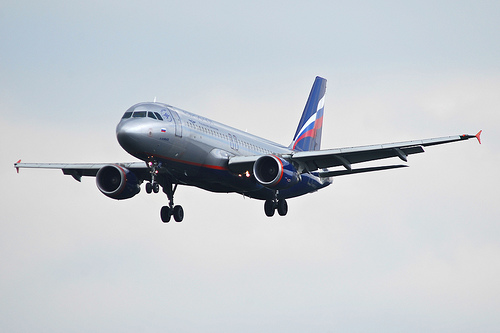} &
        \includegraphics[width=.08\linewidth]{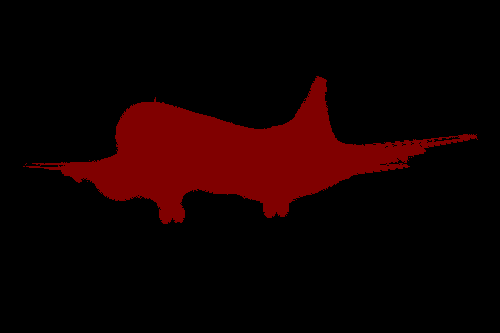} &
        \includegraphics[width=.08\linewidth]{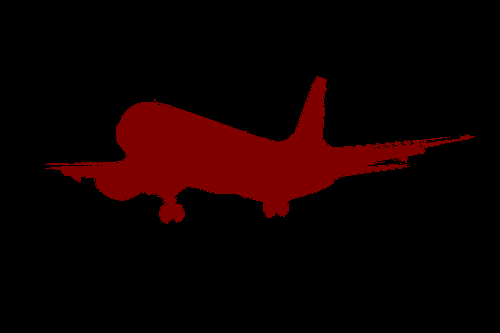} &
        \includegraphics[width=.08\linewidth]{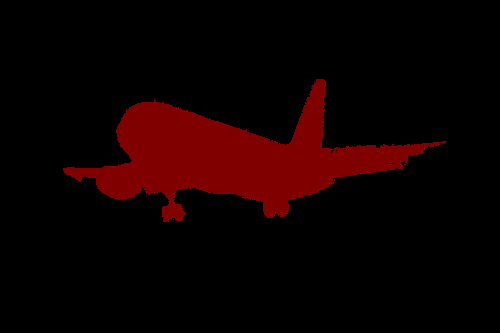} &
        \includegraphics[width=.08\linewidth]{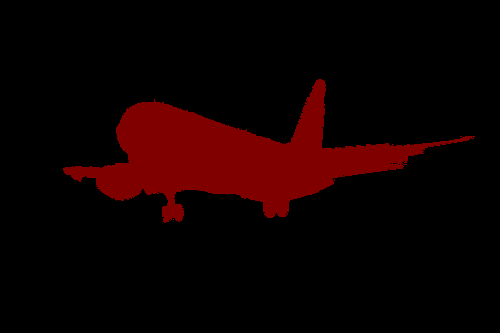} &
        \includegraphics[width=.08\linewidth]{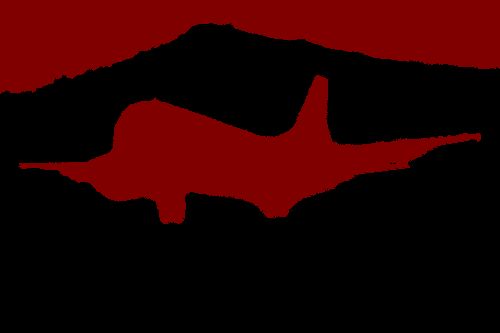} &
        \includegraphics[width=.08\linewidth]{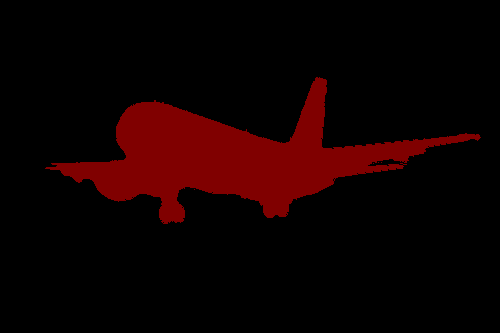} &
        \includegraphics[width=.08\linewidth]{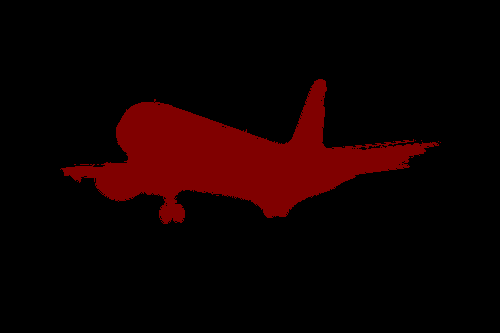} &
        \includegraphics[width=.08\linewidth]{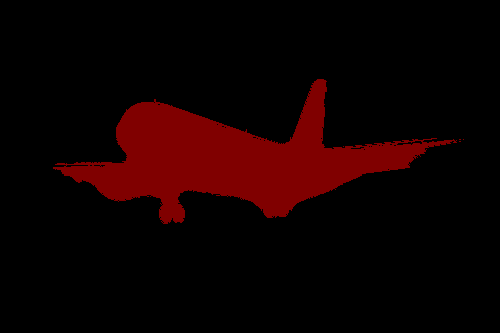} &
        \includegraphics[width=.08\linewidth]{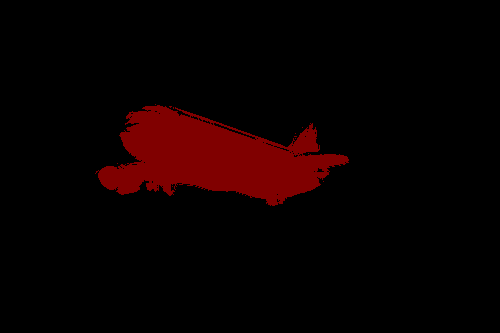} &
        \includegraphics[width=.08\linewidth]{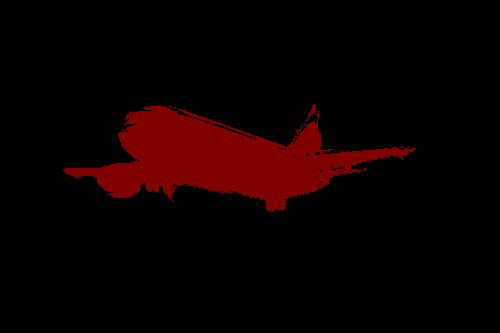} &
        \includegraphics[width=.08\linewidth]{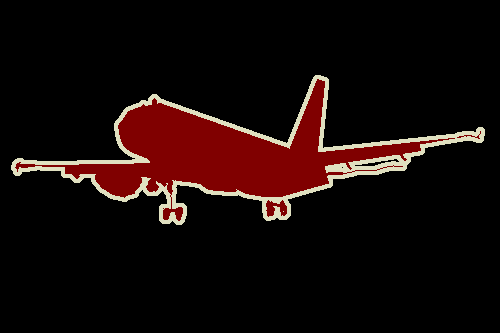} \\[2pt]
        Image & SEAM & SEAM++ & SIPE & SIPE++ & PMM & PMM++ & MCT & MCT++ & S-BCE & S-BCE++ & GT
    \end{tabular}
    \vspace{-4pt}
    \caption{Qualitative results for the implemented methods on VOC. ``++'' indicates that ISL and FSL were used for training.}
    \label{fig_qualitative}
    \vspace{-8pt}
\end{figure*}

\textbf{Binomial versus multinomial.} To isolate the effect of modeling the binomial posterior versus the multinomial posterior, we evaluate the baseline method SEAM using both, and apply ISL and FSL separately. The segmentation performance is displayed in Table \ref{table_ablation_likelihood_vs_posterior}. Both models are trained with their respective optimal values for $\lambda$, which are 0.2 for the binomial, and 0.6 for the multinomial, when used with GAP. The background parameter $\alpha$, for amplifying and weakening the class scores during CRF for AffinityNet, are also chosen to suit the respective models. 1 and 24 are used for the binomial, and 2 and 4 for the multinomial. With ISL, modeling the binomial improves the region similarity by $+1.3$ points over the multinomial, and the contour quality is improved by $+1.2$ points. With FSL, training becomes unstable when modeling the multinomial. We observed a big variance over five runs, with the region similarity ranging from 40 to 61. However, it is worth noting that for successful runs, the contour quality reached as far as 48. But on average, modeling the multinomial underperformed the binomial. Note that, not only is the performance improved with the binomial. A further benefit is that it allows ISL and FSL to be applied seamlessly to essentially any method.

\textbf{Number of sampled pixels for ISL.} In \cref{table_ablation_samples} we show segmentation results when varying the number of sampled pixels during importance sampling. We observe a boost in performance when increasing the number of samples up to 10, after which it plateaus. Thus, we fix the number of samples to 10, in order to keep the computational cost as low as possible while maintaining a high segmentation performance. A likely reason for why multiple samples improve results is that individual samples have less of an impact. This is vital for the early parts of training where predictions are mostly random. A single poor sample, such as a misclassified foreground pixel, is less likely to hamper learning when multiple samples are drawn. This effect diminishes beyond a certain point, as can be seen in \cref{table_ablation_samples}.

\begin{table}[t]
    \small
    \renewcommand{\arraystretch}{0.9}
    \setlength{\tabcolsep}{5pt}
    \begin{center}
        \caption{Region similarity ($\mathcal{J}$), contour quality ($\mathcal{F}$), and averaged ($\mathcal{J}$\&$\mathcal{F}$) on the VOC training set when varying the number of sampled pixels, with $\lambda = 1$ and without FSL.}
        \vspace{-6pt}
	\label{table_ablation_samples}
	\begin{tabular}{lccc}
		\toprule
		\#pixels & $\mathcal{J}$ & $\mathcal{F}$ & $\mathcal{J}$\&$\mathcal{F}$ \\
		\midrule
            1 & 61.6{\tiny $\pm$0.8} & 40.4{\tiny $\pm$0.6} & 51.0{\tiny $\pm$0.7} \\
            5 & 62.6{\tiny $\pm$0.6} & 40.7{\tiny $\pm$0.6} & 51.6{\tiny $\pm$0.6} \\
            10 & \textbf{63.0}{\tiny $\pm$0.4} & 41.2{\tiny $\pm$0.4} & \textbf{52.1}{\tiny $\pm$0.4} \\
            50 & \textbf{63.0}{\tiny $\pm$0.6} & \textbf{41.3}{\tiny $\pm$0.7} & \textbf{52.1}{\tiny $\pm$0.6} \\
            100 & 62.9{\tiny $\pm$0.7} & 41.2{\tiny $\pm$0.7} & 52.0{\tiny $\pm$0.7} \\
		\bottomrule
	\end{tabular}
    \end{center}
    \vspace{-16pt}
\end{table}

\begin{figure}[t]
    \centering
    \includegraphics[width=0.84\linewidth]{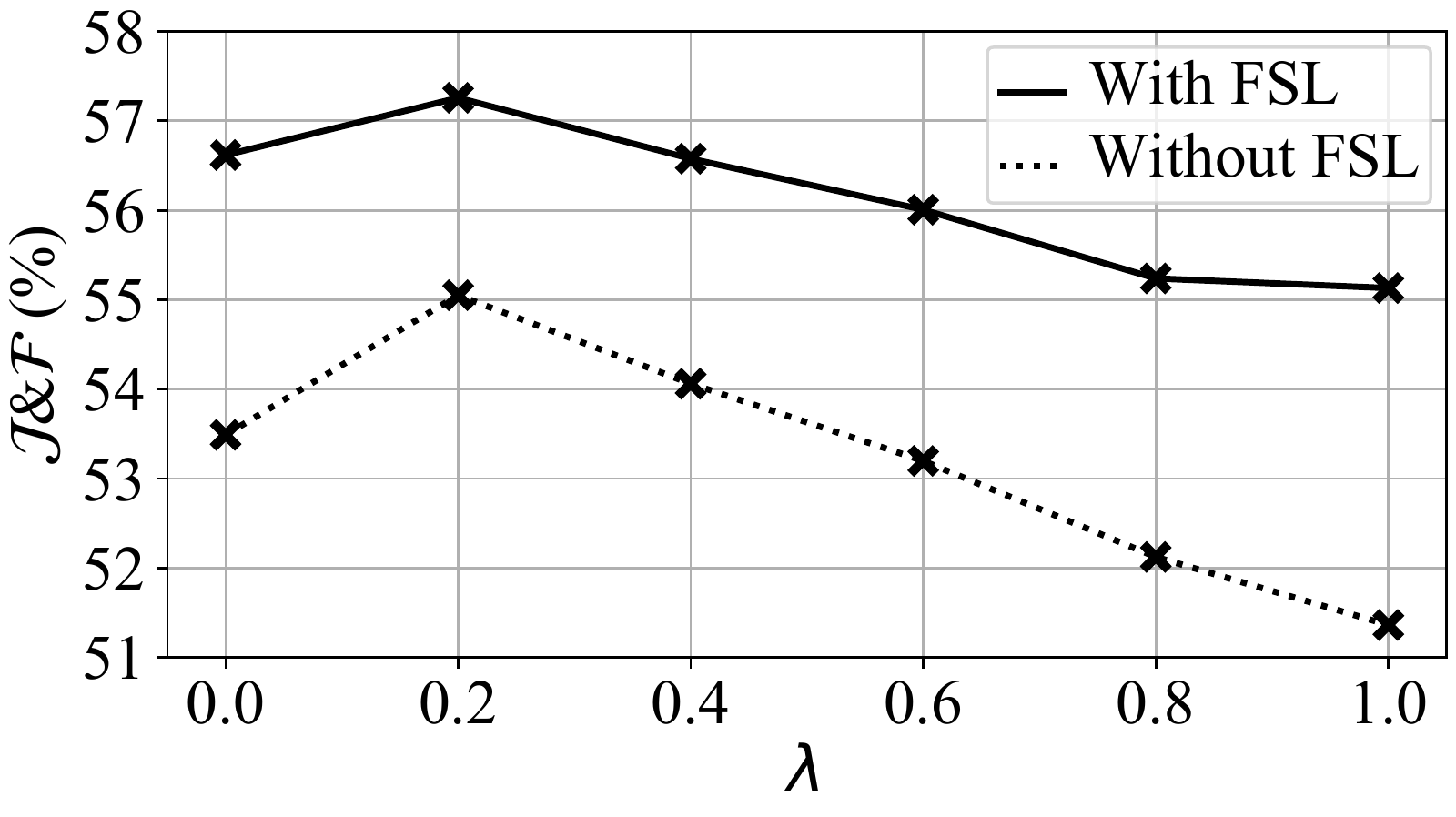}
    \vspace{-8pt}
    \caption{Combined score ($\mathcal{J}\&\mathcal{F}$) on the VOC training set, as a function of the importance sampling loss parameter $\lambda$, with and without feature similarity loss (FSL), with $\mu=2.5$ and $\sigma=5$.}
    \label{fig_ablation_lambda}
    \vspace{-6pt}
\end{figure}

\textbf{Optimal loss weight $\lambda$ for ISL.} \Cref{fig_ablation_lambda} shows the segmentation performance when varying the loss weight $\lambda$ in \eqref{eq_cls_loss}. Note that $\lambda=0$ corresponds to the standard GAP-based classification loss in \eqref{eq_ce_loss} without ISL, while $\lambda>0$ uses ISL. ISL improves over the standard loss by $+1.6$ points with $\lambda=0.2$. FSL further boosts the performance by $+2.2$ points. ISL had a larger impact on the contour quality, while FSL improved both metrics roughly the same.

\textbf{Gating function in FSL.} We compare $g(r_i, r_j)$ based on the max-normalized CAM $r$ in \eqref{eq_relu_maxnorm}, and $g(b_i, b_j)$ based on the posterior estimate $b$ in \eqref{eq_cam_sigmoid}. The max-normalized CAM, $r$, performs better with $\mathcal{J}$=66.9 (\textit{cf.}\ 64.6), $\mathcal{F}$=43.5 (\textit{cf.}\ 43.0), and $\mathcal{J}\&\mathcal{F}$=55.2 (\textit{cf.}\ 53.8). This could be attributed to the fact that the gradient is suppressed too much for $b$, which is upper bounded by $\exp (-x)$, and diminishes faster than $1/x$ for $r$. See \cref{sec_likelihood_fsl}. Additionally, $g(r_i, r_j)$ induces a lower score bound for when FSL is applied. This is not the case for $g(b_i, b_j)$, which may result in artificially large negative scores that are not reflective of the true class probability.

\section{Conclusions}

In this work, we generalize two previous techniques, importance sampling, and feature similarity loss, for weakly-supervised segmentation. Based on the argument that object classes are not mutually exclusive within image regions, we model multiple binary problems as opposed to the multinomial posterior. This results in an add-on method that boosts pseudo-label fidelity, which in turn improves segmentation performance, as shown on several state-of-the-art baselines.

\section*{Acknowledgments}

This work was partially supported by the Wallenberg AI, Autonomous Systems and Software Program (WASP), funded by the Knut and Alice Wallenberg (KAW) Foundation. The computational resources were provided by the National Academic Infrastructure for Supercomputing in Sweden (NAISS), partially funded by the Swedish Research Council through grant agreement no.\ 2022-06725, and by the Berzelius resource, provided by the KAW Foundation at the National Supercomputer Centre (NSC).

{\small
\bibliographystyle{ieee_fullname}
\bibliography{egbib}

\begin{thebibliography}{10}\itemsep=-1pt

\bibitem{ahn2019cvpr}
Jiwoon Ahn, Sunghyun Cho, and Suha Kwak.
\newblock Weakly supervised learning of instance segmentation with inter-pixel relations.
\newblock In {\em Proceedings of the IEEE/CVF Conference on Computer Vision and Pattern Recognition}, pages 2209--2218, 2019.

\bibitem{ahn2018cvpr}
Jiwoon Ahn and Suha Kwak.
\newblock Learning pixel-level semantic affinity with image-level supervision for weakly supervised semantic segmentation.
\newblock In {\em Proceedings of the IEEE Conference on Computer Vision and Pattern Recognition}, pages 4981--4990, 2018.

\bibitem{aimar2023balanced}
Emanuel~Sanchez Aimar, Arvi Jonnarth, Michael Felsberg, and Marco Kuhlmann.
\newblock Balanced product of calibrated experts for long-tailed recognition.
\newblock In {\em Proceedings of the IEEE/CVF Conference on Computer Vision and Pattern Recognition (CVPR)}, pages 19967--19977, 2023.

\bibitem{bearman2016s}
Amy Bearman, Olga Russakovsky, Vittorio Ferrari, and Li Fei-Fei.
\newblock What’s the point: Semantic segmentation with point supervision.
\newblock In {\em Computer Vision--ECCV 2016: 14th European Conference, Amsterdam, The Netherlands, October 11--14, 2016, Proceedings, Part VII 14}, pages 549--565. Springer, 2016.

\bibitem{bearman2016eccv}
Amy Bearman, Olga Russakovsky, Vittorio Ferrari, and Li Fei-Fei.
\newblock What’s the point: Semantic segmentation with point supervision.
\newblock In {\em European Conference on Computer Vision}, pages 549--565. Springer, 2016.

\bibitem{chen2020weakly}
Liyi Chen, Weiwei Wu, Chenchen Fu, Xiao Han, and Yuntao Zhang.
\newblock Weakly supervised semantic segmentation with boundary exploration.
\newblock In {\em Computer Vision--ECCV 2020: 16th European Conference, Glasgow, UK, August 23--28, 2020, Proceedings, Part XXVI 16}, pages 347--362. Springer, 2020.

\bibitem{qichen2022cvpr}
Qi Chen, Lingxiao Yang, Jian-Huang Lai, and Xiaohua Xie.
\newblock Self-supervised image-specific prototype exploration for weakly supervised semantic segmentation.
\newblock In {\em Proceedings of the IEEE/CVF Conference on Computer Vision and Pattern Recognition (CVPR)}, pages 4288--4298, June 2022.

\bibitem{zhaozhengchen2022cvpr}
Zhaozheng Chen, Tan Wang, Xiongwei Wu, Xian-Sheng Hua, Hanwang Zhang, and Qianru Sun.
\newblock Class re-activation maps for weakly-supervised semantic segmentation.
\newblock In {\em Proceedings of the IEEE/CVF Conference on Computer Vision and Pattern Recognition (CVPR)}, pages 969--978, June 2022.

\bibitem{cordts2016cityscapes}
Marius Cordts, Mohamed Omran, Sebastian Ramos, Timo Rehfeld, Markus Enzweiler, Rodrigo Benenson, Uwe Franke, Stefan Roth, and Bernt Schiele.
\newblock The cityscapes dataset for semantic urban scene understanding.
\newblock In {\em Proceedings of the IEEE conference on computer vision and pattern recognition}, pages 3213--3223, 2016.

\bibitem{dai2015iccv}
Jifeng Dai, Kaiming He, and Jian Sun.
\newblock Boxsup: Exploiting bounding boxes to supervise convolutional networks for semantic segmentation.
\newblock In {\em Proceedings of the IEEE International Conference on Computer Vision}, pages 1635--1643, 2015.

\bibitem{du2022cvpr}
Ye Du, Zehua Fu, Qingjie Liu, and Yunhong Wang.
\newblock Weakly supervised semantic segmentation by pixel-to-prototype contrast.
\newblock In {\em Proceedings of the IEEE/CVF Conference on Computer Vision and Pattern Recognition (CVPR)}, pages 4320--4329, June 2022.

\bibitem{everingham2010ijvc}
Mark Everingham, Luc Van~Gool, Christopher~KI Williams, John Winn, and Andrew Zisserman.
\newblock The pascal visual object classes (voc) challenge.
\newblock {\em International Journal of Computer Vision}, 88(2):303--338, 2010.

\bibitem{fan2020cvpr}
Junsong Fan, Zhaoxiang Zhang, Chunfeng Song, and Tieniu Tan.
\newblock Learning integral objects with intra-class discriminator for weakly-supervised semantic segmentation.
\newblock In {\em Proceedings of the IEEE/CVF Conference on Computer Vision and Pattern Recognition}, pages 4283--4292, 2020.

\bibitem{fan2020aaai}
Junsong Fan, Zhaoxiang Zhang, Tieniu Tan, Chunfeng Song, and Jun Xiao.
\newblock Cian: Cross-image affinity net for weakly supervised semantic segmentation.
\newblock In {\em Proceedings of the AAAI Conference on Artificial Intelligence}, volume~34, pages 10762--10769, 2020.

\bibitem{gruosso2021human}
Monica Gruosso, Nicola Capece, and Ugo Erra.
\newblock Human segmentation in surveillance video with deep learning.
\newblock {\em Multimedia Tools and Applications}, 80:1175--1199, 2021.

\bibitem{hariharan2011iccv}
Bharath Hariharan, Pablo Arbel{\'a}ez, Lubomir Bourdev, Subhransu Maji, and Jitendra Malik.
\newblock Semantic contours from inverse detectors.
\newblock In {\em 2011 International Conference on Computer Vision}, pages 991--998. IEEE, 2011.

\bibitem{huang2018cvpr}
Zilong Huang, Xinggang Wang, Jiasi Wang, Wenyu Liu, and Jingdong Wang.
\newblock Weakly-supervised semantic segmentation network with deep seeded region growing.
\newblock In {\em Proceedings of the IEEE Conference on Computer Vision and Pattern Recognition}, pages 7014--7023, 2018.

\bibitem{jaccard1901svsn}
Paul Jaccard.
\newblock Distribution de la flore alpine dans le bassin des dranses et dans quelques r{\'e}gions voisines.
\newblock {\em Bulletin de la Soci{\'e}t{\'e} Vaudoise des Sciences Naturelles}, 37:241--272, 1901.

\bibitem{jonnarth2022icassp}
Arvi Jonnarth and Michael Felsberg.
\newblock Importance sampling {CAM}s for weakly-supervised segmentation.
\newblock In {\em IEEE International Conference on Acoustics, Speech, and Signal Processing (ICASSP)}, pages 2639--2643, 2022.

\bibitem{khoreva2017cvpr}
Anna Khoreva, Rodrigo Benenson, Jan Hosang, Matthias Hein, and Bernt Schiele.
\newblock Simple does it: Weakly supervised instance and semantic segmentation.
\newblock In {\em Proceedings of the IEEE Conference on Computer Vision and Pattern Recognition}, pages 876--885, 2017.

\bibitem{kolesnikov2016eccv}
Alexander Kolesnikov and Christoph~H Lampert.
\newblock Seed, expand and constrain: Three principles for weakly-supervised image segmentation.
\newblock In {\em European Conference on Computer Vision}, pages 695--711. Springer, 2016.

\bibitem{krahenbuhl2011neurips}
Philipp Kr{\"a}henb{\"u}hl and Vladlen Koltun.
\newblock Efficient inference in fully connected crfs with gaussian edge potentials.
\newblock {\em Advances in Neural Information Processing Systems}, 24:109--117, 2011.

\bibitem{kweon2021iccv}
Hyeokjun Kweon, Sung-Hoon Yoon, Hyeonseong Kim, Daehee Park, and Kuk-Jin Yoon.
\newblock Unlocking the potential of ordinary classifier: Class-specific adversarial erasing framework for weakly supervised semantic segmentation.
\newblock In {\em Proceedings of the IEEE/CVF International Conference on Computer Vision (ICCV)}, pages 6994--7003, October 2021.

\bibitem{lee2021neurips}
Jungbeom Lee, Jooyoung Choi, Jisoo Mok, and Sungroh Yoon.
\newblock Reducing information bottleneck for weakly supervised semantic segmentation.
\newblock {\em Advances in Neural Information Processing Systems}, 34, 2021.

\bibitem{lee2019cvpr}
Jungbeom Lee, Eunji Kim, Sungmin Lee, Jangho Lee, and Sungroh Yoon.
\newblock Ficklenet: Weakly and semi-supervised semantic image segmentation using stochastic inference.
\newblock In {\em Proceedings of the IEEE/CVF Conference on Computer Vision and Pattern Recognition}, pages 5267--5276, 2019.

\bibitem{lee2021anti}
Jungbeom Lee, Eunji Kim, and Sungroh Yoon.
\newblock Anti-adversarially manipulated attributions for weakly and semi-supervised semantic segmentation.
\newblock In {\em Proceedings of the IEEE/CVF Conference on Computer Vision and Pattern Recognition}, pages 4071--4080, 2021.

\bibitem{lee2021bbam}
Jungbeom Lee, Jihun Yi, Chaehun Shin, and Sungroh Yoon.
\newblock Bbam: Bounding box attribution map for weakly supervised semantic and instance segmentation.
\newblock In {\em Proceedings of the IEEE/CVF conference on computer vision and pattern recognition}, pages 2643--2652, 2021.

\bibitem{li2022cvpr}
Jing Li, Junsong Fan, and Zhaoxiang Zhang.
\newblock Towards noiseless object contours for weakly supervised semantic segmentation.
\newblock In {\em Proceedings of the IEEE/CVF Conference on Computer Vision and Pattern Recognition (CVPR)}, pages 16856--16865, June 2022.

\bibitem{li2021iccv}
Yi Li, Zhanghui Kuang, Liyang Liu, Yimin Chen, and Wayne Zhang.
\newblock Pseudo-mask matters in weakly-supervised semantic segmentation.
\newblock In {\em Proceedings of the IEEE/CVF International Conference on Computer Vision (ICCV)}, pages 6964--6973, October 2021.

\bibitem{lin2016scribblesup}
Di Lin, Jifeng Dai, Jiaya Jia, Kaiming He, and Jian Sun.
\newblock Scribblesup: Scribble-supervised convolutional networks for semantic segmentation.
\newblock In {\em Proceedings of the IEEE conference on computer vision and pattern recognition}, pages 3159--3167, 2016.

\bibitem{lin2014eccv}
Tsung-Yi Lin, Michael Maire, Serge Belongie, James Hays, Pietro Perona, Deva Ramanan, Piotr Doll{\'a}r, and C~Lawrence Zitnick.
\newblock Microsoft coco: Common objects in context.
\newblock In {\em European Conference on Computer Vision}, pages 740--755. Springer, 2014.

\bibitem{loshchilov2019iclr}
Ilya Loshchilov and Frank Hutter.
\newblock Decoupled weight decay regularization.
\newblock In {\em International Conference on Learning Representations}, 2019.

\bibitem{menon2020adjustment}
Aditya~Krishna Menon, Sadeep Jayasumana, Ankit~Singh Rawat, Himanshu Jain, Andreas Veit, and Sanjiv Kumar.
\newblock Long-tail learning via logit adjustment.
\newblock In {\em International Conference on Learning Representations (ICLR)}, 2021.

\bibitem{minaee2021image}
Shervin Minaee, Yuri~Y Boykov, Fatih Porikli, Antonio~J Plaza, Nasser Kehtarnavaz, and Demetri Terzopoulos.
\newblock Image segmentation using deep learning: A survey.
\newblock {\em IEEE transactions on pattern analysis and machine intelligence}, 2021.

\bibitem{papandreou2015iccv}
George Papandreou, Liang-Chieh Chen, Kevin~P Murphy, and Alan~L Yuille.
\newblock Weakly- and semi-supervised learning of a deep convolutional network for semantic image segmentation.
\newblock In {\em Proceedings of the IEEE International Conference on Computer Vision}, pages 1742--1750, 2015.

\bibitem{pathak2015iccv}
Deepak Pathak, Philipp Krähenbühl, and Trevor Darrell.
\newblock Constrained convolutional neural networks for weakly supervised segmentation.
\newblock In {\em Proceedings of the IEEE International Conference on Computer Vision}, pages 1796--1804, 2015.

\bibitem{perazzi2016cvpr}
Federico Perazzi, Jordi Pont-Tuset, Brian McWilliams, Luc Van~Gool, Markus Gross, and Alexander Sorkine-Hornung.
\newblock A benchmark dataset and evaluation methodology for video object segmentation.
\newblock In {\em Proceedings of the IEEE Conference on Computer Vision and Pattern Recognition}, pages 724--732, 2016.

\bibitem{qi2016eccv}
Xiaojuan Qi, Zhengzhe Liu, Jianping Shi, Hengshuang Zhao, and Jiaya Jia.
\newblock Augmented feedback in semantic segmentation under image level supervision.
\newblock In {\em European Conference on Computer Vision}, pages 90--105. Springer, 2016.

\bibitem{rossetti2022eccv}
Simone Rossetti, Damiano Zappia, Marta Sanzari, Marco Schaerf, and Fiora Pirri.
\newblock Max pooling with vision transformers reconciles class and shape in weakly supervised semantic segmentation.
\newblock In {\em Computer Vision--ECCV 2022: 17th European Conference, Tel Aviv, Israel, October 23--27, 2022, Proceedings, Part XXX}, pages 446--463. Springer, 2022.

\bibitem{ru2022cvpr}
Lixiang Ru, Yibing Zhan, Baosheng Yu, and Bo Du.
\newblock Learning affinity from attention: End-to-end weakly-supervised semantic segmentation with transformers.
\newblock In {\em Proceedings of the IEEE/CVF Conference on Computer Vision and Pattern Recognition (CVPR)}, pages 16846--16855, June 2022.

\bibitem{shimoda2019iccv}
Wataru Shimoda and Keiji Yanai.
\newblock Self-supervised difference detection for weakly-supervised semantic segmentation.
\newblock In {\em Proceedings of the IEEE/CVF International Conference on Computer Vision}, pages 5208--5217, 2019.

\bibitem{su2021iccv}
Yukun Su, Ruizhou Sun, Guosheng Lin, and Qingyao Wu.
\newblock Context decoupling augmentation for weakly supervised semantic segmentation.
\newblock In {\em Proceedings of the IEEE/CVF International Conference on Computer Vision (ICCV)}, pages 7004--7014, October 2021.

\bibitem{sun2020eccv}
Guolei Sun, Wenguan Wang, Jifeng Dai, and Luc Van~Gool.
\newblock Mining cross-image semantics for weakly supervised semantic segmentation.
\newblock In {\em European conference on computer vision}, pages 347--365. Springer, 2020.

\bibitem{sun2021iccv}
Kunyang Sun, Haoqing Shi, Zhengming Zhang, and Yongming Huang.
\newblock Ecs-net: Improving weakly supervised semantic segmentation by using connections between class activation maps.
\newblock In {\em Proceedings of the IEEE/CVF International Conference on Computer Vision (ICCV)}, pages 7283--7292, October 2021.

\bibitem{vernaza2017cvpr}
Paul Vernaza and Manmohan Chandraker.
\newblock Learning random-walk label propagation for weakly-supervised semantic segmentation.
\newblock In {\em Proceedings of the IEEE Conference on Computer Vision and Pattern Recognition}, pages 7158--7166, 2017.

\bibitem{wang2020cvpr}
Yude Wang, Jie Zhang, Meina Kan, Shiguang Shan, and Xilin Chen.
\newblock Self-supervised equivariant attention mechanism for weakly supervised semantic segmentation.
\newblock In {\em Proceedings of the IEEE/CVF Conference on Computer Vision and Pattern Recognition}, pages 12275--12284, 2020.

\bibitem{wei2017cvpr}
Yunchao Wei, Jiashi Feng, Xiaodan Liang, Ming-Ming Cheng, Yao Zhao, and Shuicheng Yan.
\newblock Object region mining with adversarial erasing: A simple classification to semantic segmentation approach.
\newblock In {\em Proceedings of the IEEE conference on computer vision and pattern recognition}, pages 1568--1576, 2017.

\bibitem{wu2022adaptive}
Tong Wu, Guangyu Gao, Junshi Huang, Xiaolin Wei, Xiaoming Wei, and Chi~Harold Liu.
\newblock Adaptive spatial-bce loss for weakly supervised semantic segmentation.
\newblock In {\em Computer Vision--ECCV 2022: 17th European Conference, Tel Aviv, Israel, October 23--27, 2022, Proceedings, Part XXIX}, pages 199--216. Springer, 2022.

\bibitem{xu2022cvpr}
Lian Xu, Wanli Ouyang, Mohammed Bennamoun, Farid Boussaid, and Dan Xu.
\newblock Multi-class token transformer for weakly supervised semantic segmentation.
\newblock In {\em Proceedings of the IEEE/CVF Conference on Computer Vision and Pattern Recognition (CVPR)}, pages 4310--4319, June 2022.

\bibitem{yoon2022eccv}
Sung-Hoon Yoon, Hyeokjun Kweon, Jegyeong Cho, Shinjeong Kim, and Kuk-Jin Yoon.
\newblock Adversarial erasing framework via triplet with gated pyramid pooling layer for weakly supervised semantic segmentation.
\newblock In {\em Computer Vision--ECCV 2022: 17th European Conference, Tel Aviv, Israel, October 23--27, 2022, Proceedings, Part XXIX}, pages 326--344. Springer, 2022.

\bibitem{zhang2020neurips}
Dong Zhang, Hanwang Zhang, Jinhui Tang, Xian-Sheng Hua, and Qianru Sun.
\newblock Causal intervention for weakly-supervised semantic segmentation.
\newblock In H. Larochelle, M. Ranzato, R. Hadsell, M.~F. Balcan, and H. Lin, editors, {\em Advances in Neural Information Processing Systems}, volume~33, pages 655--666. Curran Associates, Inc., 2020.

\bibitem{zhang2021iccv}
Fei Zhang, Chaochen Gu, Chenyue Zhang, and Yuchao Dai.
\newblock Complementary patch for weakly supervised semantic segmentation.
\newblock In {\em Proceedings of the IEEE/CVF International Conference on Computer Vision}, pages 7242--7251, 2021.

\bibitem{zhou2016cvpr}
Bolei Zhou, Aditya Khosla, Agata Lapedriza, Aude Oliva, and Antonio Torralba.
\newblock Learning deep features for discriminative localization.
\newblock In {\em Proceedings of the IEEE Conference on Computer Vision and Pattern Recognition}, pages 2921--2929, 2016.

\end{thebibliography}
}


\twocolumn[\vspace{40pt}{\centering{\fontsize{14}{17}\selectfont \textbf{Supplementary Material}\par}\vspace{40pt}}]
\appendix
\setcounter{section}{19}

This supplementary material contains three parts. (1) The ablation study on how we find the optimal values for the parameters $\mu$ and $\sigma$ in FSL in Sec.\ \ref{sec_ablation_mu_sigma}. (2) An extended state-of-the-art comparison on VOC in Sec.\ \ref{sec_voc_test}. (3) A discussion on reproducibility in weakly-supervised semantic segmentation in Sec.\ \ref{sec_reproducibility}.

\subsection{Ablation study for $\mu$ and $\sigma$ in FSL}
\label{sec_ablation_mu_sigma}

The spatial extent of the feature similarity loss (FSL) is controlled by $\sigma$, while the dissimilarity threshold between pixel values is controlled by $\mu$. Since learning them together with the network parameters could lead to trivial solutions, we set them as fixed parameters and find optimal values based on the segmentation performance. In the spirit of the weakly supervised setting, we only use a handful of images, to reduce the required ground-truth masks, and randomly sample one image per class from the VOC training set. Subsequently, we optimize FSL with respect to initial CAMs, and compute the resulting region similarity, $\mathcal{J}$, and contour quality, $\mathcal{F}$. Note that no network was involved at this stage.

\begin{figure}[t]
    \centering
    \subfloat[$\sigma=5.0$]{\includegraphics[width=0.9\linewidth]{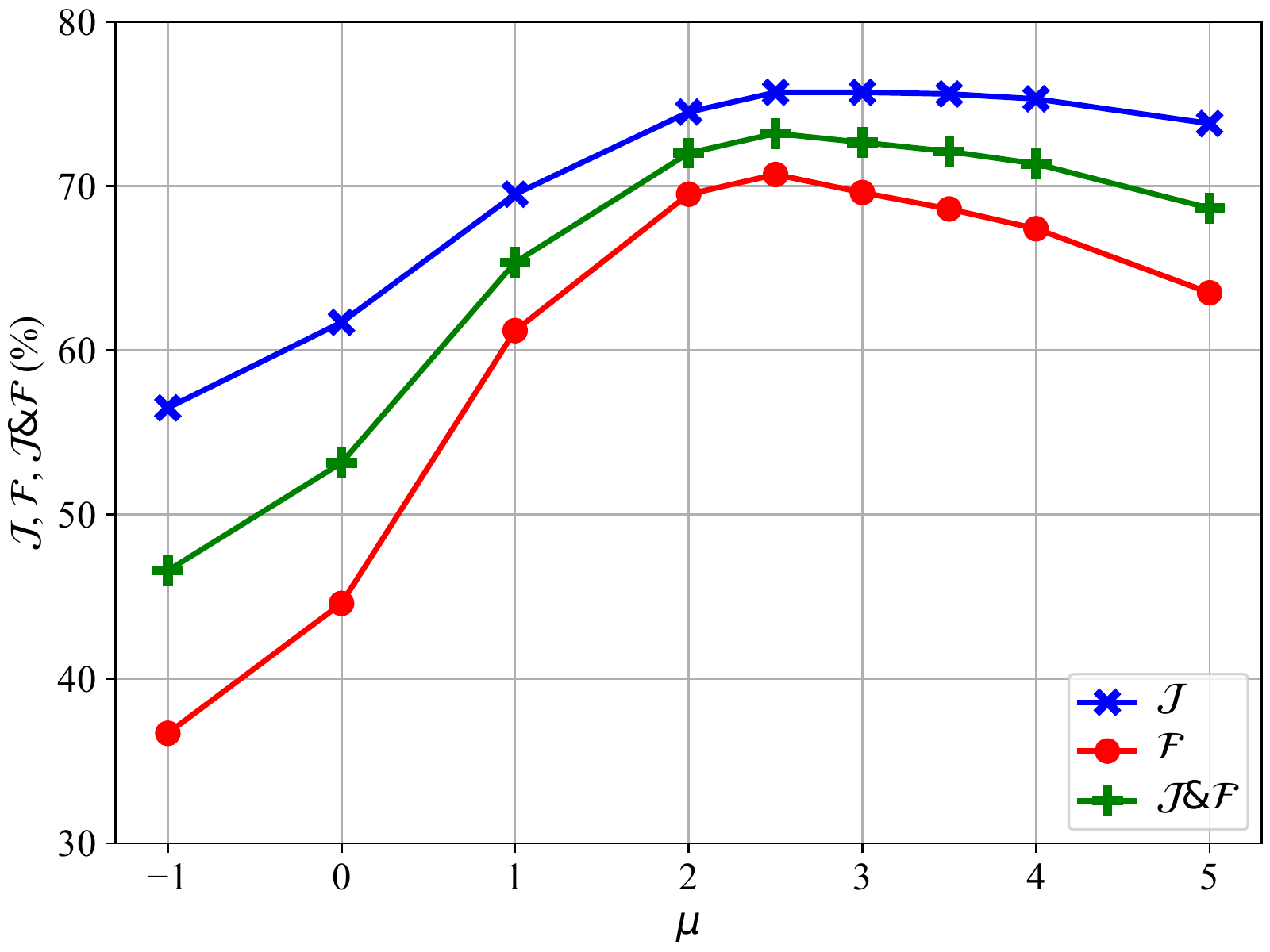}%
    \label{fig_ablation_mu}}
    
    \subfloat[$\mu=2.5$]{\includegraphics[width=0.9\linewidth]{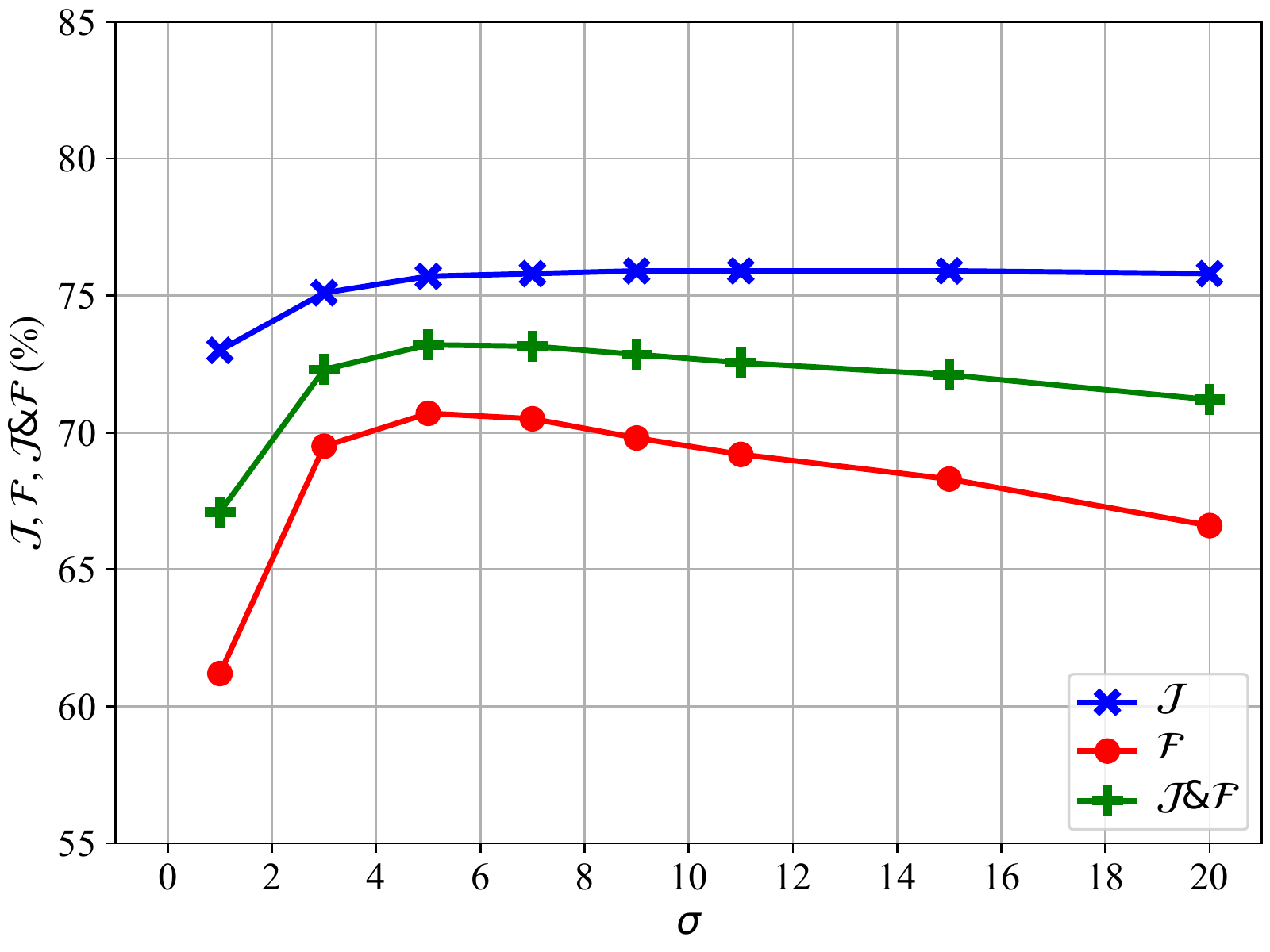}%
    \label{fig_ablation_sigma}}
    
    \caption{Region similarity ($\mathcal{J}$), contour quality ($\mathcal{F}$), and averaged ($\mathcal{J}\&\mathcal{F}$), as functions of (a) $\mu$, and (b) $\sigma$, when optimizing Gaussian CAMs on one image per class.}
    \label{fig_ablation_mu_sigma}
\end{figure}

Due to both simplicity, and to mimic realistic CAMs, we hand-craft initial CAMs for each image, based on the Gaussian function. See \cref{fig_fsl_demo} in the main paper to get an idea of what the initial CAMs look like. For the unimodal Gaussian CAMs to make sense, we only sampled images that contain a single object. The Gaussian CAMs were defined by the mean and standard deviation of the ground-truth segmentation masks. The mean was computed as the average position of all foreground pixels, and the two non-zero components of the linearly independent diagonal covariance matrix were computed as the variance in the two spatial coordinates. Finally, the non-normalized scores, or logits, were computed as
\begin{equation}
    s(i, j) = 2 G(i, j) - 1,
\end{equation}
where $G(i, j)$ is the Gaussian function at spatial location $(i, j)$. This was done to attain negative values outside the object boundaries. Note that, since we only used images with a single object, $s$ contains a single channel, and essentially predicts foreground versus background. The predicted foreground segmentation mask was attained by thresholding the score at zero, where positive values were predicted as foreground.

The segmentation performance after optimizing FSL for different values of $\mu$ and $\sigma$ are shown in \cref{fig_ablation_mu_sigma}. The highest combined score $\mathcal{J}\&\mathcal{F}$ is achieved for $\mu = 2.5$ and $\sigma = 5$, so we fix the parameters to these values.

A Gaussian spatial weight with $\sigma = 5$ means that $\sim$$39\%$ of the loss contribution comes from pixel pairs that are within 5 pixels apart, while pairs that are within 10 pixels contribute to $\sim$$86\%$ of the loss. Note that the CAM resolution is $56 \times 56$ when using the ResNet-38 backbone with an input size of $448 \times 448$, which is the case for our SEAM \cite{wang2020cvpr} baseline. Thus, the FSL optimization procedure described above was done in a tenth of the original image resolution, to roughly match the CAM resolution.

A dissimilarity threshold of $\mu = 2.5$ means that pixels are considered similar if their normalized L1 distance $\delta$ between the RGB color values in \eqref{eq_fsl_delta} of the main paper is less than $0.076$, since this corresponds to a negative pixel dissimilarity score, \ie $f(\delta) < 0$ in \eqref{eq_fsl_f}.

\subsection{Extended VOC comparison}
\label{sec_voc_test}

For further insights, \cref{table_seam_ext} shows our reproduced final segmentation results when applying ISL and FSL separately to the implemented baselines. A general trend is that ISL mainly improves the contour quality, $\mathcal{F}$, increasing it by $+1.0$ points on average, while improving the region similarity, $\mathcal{J}$, by $+0.3$ points. The feature similarity loss significantly improves both metrics, where $\mathcal{J}$ and $\mathcal{F}$ are increased by $+1.1$ and $+1.8$ points respectively.

For completeness and transparency, we show in \cref{table_sota_voc} both the reported and reproduced final segmentation results for the implemented methods. We also include reported results for other methods that we did not reimplement. However, methods that use additional supervision, either directly using other datasets, or indirectly using saliency maps, are excluded. The reported results are taken directly from the respective publications, while our reproduced results on the validation set are computed as the average over five runs, and may thus deviate. For the test set, we submit the segmentation predictions from the best out of the five runs, based on validation set performance, to the PASCAL VOC evaluation server.

Comparing our reproduced results, our proposed losses improve $\mathcal{J}$ on the validation set by +1.5, and on the test set by +1.1 points on average.

Compared to the reported results, we improve $\mathcal{J}$ for SEAM \cite{wang2020cvpr}, and SIPE \cite{qichen2022cvpr} using ResNet-101, by +1.9 and +0.6 respectively on the test set. Since we did not manage to reproduce the other methods fully, their test scores were not improved compared to the reported results. See Sec.\ \ref{sec_reproducibility} for potential reasons for this.

\begin{table}[t]
    \renewcommand{\arraystretch}{0.9}
    \setlength{\tabcolsep}{3pt}
    \begin{center}
        \caption{Final segmentation performance on the VOC validation set, comparing ISL and FSL separately on different state-of-the-art baselines in terms of region similarity ($\mathcal{J}$), contour quality ($\mathcal{F}$), and combined ($\mathcal{J}\&\mathcal{F}$).}
        \label{table_seam_ext}
        \begin{tabular}{llccccc}
            \toprule
            Method & Backb. & ISL & FSL & $\mathcal{J}$ & $\mathcal{F}$ & $\mathcal{J}\&\mathcal{F}$ \\
            \midrule
            SEAM & Res38 & & & 63.9{\tiny $\pm$0.5} & 39.9{\tiny $\pm$0.2} & 51.9{\tiny $\pm$0.3} \\
            \cite{wang2020cvpr} & Res38 & \checkmark & & 64.3{\tiny $\pm$0.8} & 42.2{\tiny $\pm$0.6} & 53.3{\tiny $\pm$0.7} \\
            & Res38 & & \checkmark & 66.9{\tiny $\pm$0.4} & 43.5{\tiny $\pm$0.2} & 55.2{\tiny $\pm$0.2} \\
            \midrule
            SIPE & Res38 & & & 68.0{\tiny $\pm$0.2} & 45.1{\tiny $\pm$0.2} & 56.6{\tiny $\pm$0.1} \\
            \cite{qichen2022cvpr} & Res38 & \checkmark & & 68.1{\tiny $\pm$0.4} & 46.2{\tiny $\pm$0.2} & 57.1{\tiny $\pm$0.2} \\
            & Res38 & & \checkmark & 68.4{\tiny $\pm$0.3} & 46.3{\tiny $\pm$0.2} & 57.4{\tiny $\pm$0.2} \\
            \midrule
            SIPE & Res101 & & & 68.5{\tiny $\pm$0.2} & 41.9{\tiny $\pm$0.5} & 55.2{\tiny $\pm$0.3} \\
            \cite{qichen2022cvpr} & Res101 & \checkmark & & 69.2{\tiny $\pm$0.2} & 43.3{\tiny $\pm$0.3} & 56.3{\tiny $\pm$0.3} \\
            & Res101 & & \checkmark & 68.9{\tiny $\pm$0.2} & 43.0{\tiny $\pm$0.2} & 56.0{\tiny $\pm$0.2} \\
            \midrule
            PMM & Res38 & & & 64.7{\tiny $\pm$0.5} & 44.5{\tiny $\pm$0.5} & 54.6{\tiny $\pm$0.4} \\
            \cite{li2021iccv} & Res38 & \checkmark & & 65.0{\tiny $\pm$1.0} & 44.6{\tiny $\pm$0.5} & 54.8{\tiny $\pm$0.5} \\
            & Res38 & & \checkmark & 66.0{\tiny $\pm$0.3} & 46.0{\tiny $\pm$0.5} & 56.0{\tiny $\pm$0.4} \\
            \midrule
            \small{MCTformer} & DeiT-S & & & 67.5{\tiny $\pm$1.7} & 46.7{\tiny $\pm$0.7} & 57.1{\tiny $\pm$1.2} \\
            \cite{xu2022cvpr} & DeiT-S & \checkmark & & 67.5{\tiny $\pm$1.4} & 47.1{\tiny $\pm$0.9} & 57.3{\tiny $\pm$1.1} \\
            & DeiT-S & & \checkmark & 68.5{\tiny $\pm$1.2} & 47.0{\tiny $\pm$0.5} & 57.8{\tiny $\pm$0.9} \\
            \midrule
            \small{Spatial-BCE} & Res38 & & & 68.1{\tiny $\pm$0.1} & 45.4{\tiny $\pm$0.1} & 56.7{\tiny $\pm$0.1} \\
            \cite{wu2022adaptive} & Res38 & \checkmark & & 68.2{\tiny $\pm$0.2} & 46.2{\tiny $\pm$0.1} & 57.2{\tiny $\pm$0.1} \\
            & Res38 & & \checkmark & 68.8{\tiny $\pm$0.1} & 47.6{\tiny $\pm$0.1} & 58.2{\tiny $\pm$0.1} \\
            \midrule
            \small{Spatial-BCE} & Res101 & & & 67.9{\tiny $\pm$0.2} & 46.1{\tiny $\pm$0.1} & 57.0{\tiny $\pm$0.1} \\
            \cite{wu2022adaptive} & Res101 & \checkmark & & 68.5{\tiny $\pm$0.2} & 47.8{\tiny $\pm$0.2} & 58.1{\tiny $\pm$0.1} \\
            & Res101 & & \checkmark & 69.0{\tiny $\pm$0.2} & 48.9{\tiny $\pm$0.2} & 59.0{\tiny $\pm$0.2} \\
            \bottomrule
        \end{tabular}
    \end{center}
\end{table}

\subsection{Discussion on the limits of reproducibility}
\label{sec_reproducibility}

As stated in the main paper, we use the implementations referenced to in the respective publications, as is, with only minor modifications described in \cref{sec_implementation_details}. Still, we did not manage to reproduce the reported results exactly, in all cases. We believe that weakly-supervised segmentation is especially tricky from a reproducibility perspective, as it involves multiple steps to arrive at the final model. In chronological order, these are: (1) Downloading data and pre-trained weights; (2) training a classification network, (3) generating CAMs; (4) optionally generating labels for a pseudo-label refinement method; (5) optionally training or applying a pseudo-label refinement method, \eg IRN \cite{ahn2019cvpr} or AffinityNet \cite{ahn2018cvpr}; (6) generating pseudo-labels; (7) training a final segmentation model; (8) optionally applying a post-processing method, \eg CRF \cite{krahenbuhl2011neurips}, and finally; (9) evaluating the final segmentation predictions.

\begin{table}[t]
    \setlength{\tabcolsep}{5pt}
    \begin{center}
    \caption{Final segmentation comparison in terms of region similarity on VOC, including both reported values in the respective publications and our reproduced results.}
    \label{table_sota_voc}
    \begin{tabular}{llcccc}
    \toprule
    & & \multicolumn{2}{c}{Reported} & \multicolumn{2}{c}{Reproduced} \\
    \cmidrule(l{\tabcolsep}r{\tabcolsep}){3-4}
    \cmidrule(l{\tabcolsep}r{\tabcolsep}){5-6}
    Method & Backb. & \textit{val} & \textit{test} & \textit{val} & \textit{test} \\
    \midrule
    CCNN \cite{pathak2015iccv}         & VGG16  & 35.3 & 35.6 & - & - \\
    EM-Adapt \cite{papandreou2015iccv} & VGG16  & 38.2 & 39.6 & - & - \\
    SEC \cite{kolesnikov2016eccv}      & VGG16  & 50.7 & 51.7 & - & - \\
    AugFeed \cite{qi2016eccv}          & VGG16  & 54.3 & 55.5 & - & - \\
    AffinityNet \cite{ahn2018cvpr}     & Res38  & 61.7 & 63.7 & - & - \\
    ICD \cite{fan2020cvpr}             & Res101 & 64.1 & 64.3 & - & - \\
    CIAN \cite{fan2020aaai}            & Res101 & 64.3 & 65.3 & - & - \\
    SSDD \cite{shimoda2019iccv}        & Res38  & 64.9 & 65.5 & - & - \\
    AFA \cite{ru2022cvpr}              & MiT-B1 & 66.0 & 66.3 & - & - \\
    CONTA \cite{zhang2020neurips}      & Res38  & 66.1 & 66.7 & - & - \\
    CDA \cite{su2021iccv}              & Res38  & 66.1 & 66.8 & - & - \\
    MCIS \cite{sun2020eccv}            & Res101 & 66.2 & 66.9 & - & - \\
    PPC \cite{du2022cvpr}              & Res38  & 67.7 & 67.4 & - & - \\
    ECS-Net \cite{sun2021iccv}         & Res38  & 66.6 & 67.6 & - & - \\
    CGNet \cite{kweon2021iccv}         & Res38  & 68.4 & 68.2 & - & - \\
    ReCAM \cite{zhaozhengchen2022cvpr} & Res101 & 68.5 & 68.4 & - & - \\
    CPN \cite{zhang2021iccv}           & Res38  & 67.8 & 68.5 & - & - \\
    RIB \cite{lee2021neurips}          & Res101 & 68.3 & 68.6 & - & - \\
    ViT-PCM \cite{rossetti2022eccv}    & ViT-B  & 70.3 & 70.9 & - & - \\
    AEFT \cite{yoon2022eccv}           & Res38  & 70.9 & 71.7 & - & - \\
    SANCE~\cite{li2022cvpr}            & Res101 & 70.9 & 72.2 & - & - \\
    \midrule
    SEAM \cite{wang2020cvpr}           & Res38  & 64.5 & 65.7 & 63.9 & 65.4 \\
    \textbf{+ISL/FSL (ours)}           & Res38  & -    & -    & 66.7 & 67.6 \\
    \midrule
    SIPE \cite{qichen2022cvpr}         & Res38  & 68.2 & 69.5 & 68.0 & 68.9 \\
    \textbf{+ISL/FSL (ours)}           & Res38  & -    & -    & 68.3 & 69.4 \\
    \midrule
    SIPE \cite{qichen2022cvpr}         & Res101 & 68.8 & 69.7 & 68.5 & 69.4 \\
    \textbf{+ISL/FSL (ours)}           & Res101 & -    & -    & 69.4 & 70.3 \\
    \midrule
    PMM \cite{li2021iccv}              & Res38  & 68.5 & 69.0 & 64.7 & 65.7 \\
    \textbf{+ISL/FSL (ours)}           & Res38  & -    & -    & 66.7 & 67.0 \\
    \midrule
    MCTformer \cite{xu2022cvpr}        & DeiT-S & 71.9 & 71.6 & 67.5 & 70.6 \\
    \textbf{+ISL/FSL (ours)}           & DeiT-S & -    & -    & 68.3 & 70.0 \\
    \midrule
    Spatial-BCE \cite{wu2022adaptive}  & Res38  & 70.0 & 71.3 & 68.1 & 68.4 \\
    \textbf{+ISL/FSL (ours)}           & Res38  & -    & -    & 69.3 & 69.4 \\
    \midrule
    Spatial-BCE \cite{wu2022adaptive}  & Res101 & -    & -    & 67.9 & 68.4 \\
    \textbf{+ISL/FSL (ours)}           & Res101 & -    & -    & 70.1 & 70.6 \\
    \bottomrule
    \end{tabular}
    \end{center}
\end{table}

\begin{table*}[t]
    \setlength{\tabcolsep}{5pt}
    \begin{center}
    \caption{List of code repositories used for reproducing the results. See Sec.\ \ref{sec_reproducibility} for what each step involves.}
    \label{table_repos}
    \begin{tabular}{lll}
    \toprule
    Method & Steps & Code repository \\
    \midrule
    \multirow{2}{*}{SEAM \cite{wang2020cvpr}} & 1-6 & \small{\url{https://github.com/YudeWang/SEAM}} \\
    & 7-9 & \small{\url{https://github.com/YudeWang/semantic-segmentation-codebase}} \\
    \midrule
    \multirow{3}{*}{SIPE \cite{qichen2022cvpr}} & 1-6 & \small{\url{https://github.com/chenqi1126/SIPE}} \\
    & 7-9 w/ Res38 & \small{\url{https://github.com/YudeWang/semantic-segmentation-codebase}} \\
    & 7-9 w/ Res101 & \small{\url{https://github.com/kazuto1011/deeplab-pytorch}} \\
    \midrule
    \multirow{2}{*}{PMM \cite{li2021iccv}} & 1-3, 6 & \small{\url{https://github.com/Eli-YiLi/PMM}} \\
    & 7-9 & \small{\url{https://github.com/Eli-YiLi/WSSS_MMSeg}} \\
    \midrule
    MCTformer \cite{xu2022cvpr} & 1-9 & \small{\url{https://github.com/xulianuwa/MCTformer}} \\
    \midrule
    \multirow{4}{*}{Spatial-BCE \cite{wu2022adaptive}} & 1-3 & \small{\url{https://github.com/allenwu97/Spatial-BCE}} \\
    & 4-6 & \small{\url{https://github.com/jiwoon-ahn/irn}} \\
    & 7-9 w/ Res38 & \small{\url{https://github.com/YudeWang/semantic-segmentation-codebase}} \\
    & 7-9 w/ Res101 & \small{\url{https://github.com/kazuto1011/deeplab-pytorch}} \\
    \bottomrule
    \end{tabular}
    \end{center}
\end{table*}

This convoluted pipeline becomes especially tricky to reproduce, due to the fact that the steps are usually split across multiple code repositories. This introduces additional possibilities for misaligned implementation details, especially if they are not fully listed. Commonly, the authors of WSSS papers provide code for steps 1-3, and the subsequent steps are typically implemented in a different code repository, and in some cases even maintained by different authors. See \cref{table_repos} for the repositories that we used in the different steps for each method, which includes a total of 9 unique code repositories. A further argument that speaks for this being the main reason, is that we did manage to reproduce the CAM results, as can be seen in \cref{table_cam} in the main paper. The region similarity of our reproduced CAM pseudo-labels matched the reported results within the margin of error, in most cases. In all cases, the CAM results were more closely reproduced than the final segmentation results. This means that the subsequent steps introduce differences in the implementation, which is reasonable as this is typically not the main focus of WSSS papers.

In the case of MCTformer \cite{xu2022cvpr}, where all stages are contained in a single repository, we still observe a discrepancy between the reported and reproduced results. Moreover, the degree of reproducibility varies between the validation and test results. The gap is by far larger on the validation results, which could possibly be caused by hyperparameter tuning on the validation data, not reflected in the repository. Additionally, this can to some extent be explained by the high variance over five runs, where the best run reproduces the reported CAM result in \cref{table_cam} of the main paper. If the subsequent steps contain a similar variance, the reported final segmentation result could be achieved as the best out of a larger number of runs.

Furthermore, while most WSSS works carefully state the training details for steps 1-3, it is next to impossible to list the full configuration of the experiments. The following additional items could potentially affect performance:
\begin{itemize}
\itemsep0em
    \item Different software configurations, \ie choice of package manager (pip versus conda), python version, python package versions, or CUDA version \etc.
    \item Different hardware configurations, \ie number of GPUs, GPU model, CPU model \etc.
    \item Different computational environments, \ie OS version, the use of containers \etc.
\end{itemize}

\end{document}